\documentclass[table]{article}
\usepackage{jfr}
\usepackage{graphicx}
\usepackage{setspace}
\usepackage{svg}
\usepackage{caption}
\usepackage{lineno}
\usepackage{todonotes}
\usepackage{svg}
\usepackage{comment}
\usepackage{adjustbox}
\usepackage{multirow}
\usepackage{multicol}
\usepackage{textcmds}
\usepackage{siunitx}
\usepackage{booktabs}
\usepackage[utf8]{inputenc}

\DeclareUnicodeCharacter{0301}{\'{e}}

\usepackage[backend=biber, sorting=nyt, maxcitenames=2, style=apa]{biblatex}
\usepackage{soul}

\newcommand*\samethanks[1][\value{footnote}]{\footnotemark[#1]}

\usepackage[noabbrev, capitalize, nameinlink]{cleveref}

\usepackage{xspace}
\newcommand*{\cf}{cf.\@\xspace}

\usepackage[left=3cm, right=3cm, top=3cm, bottom=3cm]{geometry}

\usepackage{xcolor}

\title{Why robots should be technical: Correcting mental models through technical architecture concepts\thanks{This work is currently under review in the Interaction Studies Journal.}}

\author{Lukas~Hindemith\thanks{L. Hindemith, A.-L. Vollmer, J. P. G\"opfert and B. Wrede were with the Institute for Cognition and Robotics (CoR-Lab) and the Center for Cognitive Interaction Technology (CITEC), Bielefeld University, 33615 Bielefeld, Germany, e-mail:  \textit{\{lhindemith,avollmer,jgoepfert,bwrede\}@techfak.uni-bielefeld.de}}\\
        \And
        Anna-Lisa~Vollmer\samethanks[1]\\
        \And
        Jan~Philip~G\"opfert\samethanks[1]\\
        \And
        Christiane~B.~Wiebel-Herboth\thanks{C.~B.~Wiebel-Herboth was with Honda Research Institute Europe, 63073 Offenbach am Main, Germany, e-mail: \textit{christiane.wiebel@honda-ri.de}}\\
        \And
        Britta~Wrede\samethanks[1]
}

\begin{document}

\maketitle

\begin{abstract}
Research in social robotics is commonly focused on designing robots that imitate human behavior. While this might increase a user's satisfaction and acceptance of robots at first glance, it does not automatically aid a non-expert user in naturally interacting with robots, and might actually hurt their ability to correctly anticipate a robot's capabilities. We argue that a faulty mental model, that the user has of the robot, is one of the main sources of confusion. In this work we investigate how communicating technical concepts of robotic systems to users affects their mental models, and how this can increase the quality of human-robot interaction. We conducted an online study and investigated possible ways of improving users' mental models. Our results underline that communicating technical concepts can form an improved mental model. Consequently, we show the importance of consciously designing robots that express their capabilities and limitations.
\end{abstract}

\section{Introduction}
In recent years, the field of robotics has been growing quickly. With vacuum cleaning robots or even humanoid robots such as Pepper \autocite{pepper}, this field is no longer limited to the industrial sector. Instead new applications in private lives emerges. This change leads to new challenges for robotic engineers and researcher. Robot's working environments are no longer static and consistent as it is in factories, where their behavior can be pre-programmed. Hence, robots need to continuously adapt to new and dynamic environments. On the other hand, interaction partners are no longer experts, but naive users who want to interact with robots in an intuitive way.

Citizens of modern societies are regularly confronted with technology, and are therefore used to a number of de facto standards with regards to interfaces and the current state of the art. At the same time, most are not specialists in specific technological fields and neither require nor posses deep knowledge about the inner workings of technological systems. We henceforth refer to them as \emph{naive users}.

This need of an intuitive human-robot interaction can be approached from different directions. In current research, this problem is often approached by developing robotic systems simulating real humans' response and behavior and for example convincing users to have emotions \parencite{breazeal2016social,vollmer2018studying}. While this approach reduces the cognitive load for users and at the same time increases the satisfaction on the user's side, it also raises problems \parencite{breazeal2016social,duffy2006fundamental,hegel2011towards}.
In limited scenarios or use-cases, if the user and the robot interact in a successful way, the human-like behavior of the robot does not cause any problems. However, today's robotic systems are still limited in their functionality and therefore cannot cover the range of human capabilities. In particular the way robots learn differs greatly from the way humans do. This often leads to errors in human-robot interaction. In such situations, naive users are prone to be unable to trace the error back to its origin. This problem is due to the user's faulty \emph{mental model} about the robot.

We understand a \emph{mental model} as a cognitive framework people use to form an internal representation of the things they interact with \parencite{staggers1993mental}. People build initial mental models of things they are not familiar with based on expectations and prior experiences. This initial mental model changes continuously based on new experiences.
Thus, convincing naive users that social robots are human-like and possess human internal mechanisms causes users to apply knowledge about interactions with humans to build their mental models of robots. While this might work to some degree, a faulty mental model will cause incomprehensibility in error situations and also cause erroneous user conduct, both eventually leading to misunderstandings and dysfunctional human-robot interaction. We see learning interactions in particular as an important use case. In cases where a user teaches new skills and knowledge to a robot, a faulty mental model can lead to severe consequences. This might cause wrong sample generation or even learning incorrect skills, which takes both further away from a common mental model. Based on this mismatch between the user's mental model and the actual functionality of the system, we argue that social robots should not simulate biological functions and behavior, but communicate true technical concepts. These concepts need to be communicated in a way that the true functionality with its limitations are revealed. At the same time, users may not be overloaded with information. With advances on robot capabilities towards human-like behavior this has to be communicated as well. Thus, research should focus on the goal of shaping more appropriate mental models about robots. an improved mental model on the user's side will reduce the number of erroneous human-robot interactions. With knowledge about the system's functioning and limitations, naive users will be more proficient in coping with errors.

We think that a new direction of research should focus on communicating insights of robots' architecture. Based on the expectation that naive users, while not technophile, are accustomed to technology to some degree, we developed two different ways to communicate insights of a robotic system. These two different ways, that complement each other, are a prior instruction and a robot feedback system as visualizations. We evaluated their influence on the user's mental model in an online survey.

The remaining parts of the paper are structured as followed. In \cref{sec:relWork} we review factors that influence a mental model and ways of communicating insights of robotic systems. Based on this, we formulate our hypotheses in \cref{sec:CaH}. Our realization of the system and the experimental design will be explained in \cref{sec:meth}. In \cref{sec:res} we present our results from the conducted online survey. Afterwards, we discuss the results and relate our findings to the hypotheses in \cref{sec:discuss}. In \cref{sec:conclusion} we summarize our work and give an outlook to future work.

\section{Related Work}
\label{sec:relWork}

\subsection{Dual Nature of Computational Artifacts: Relevance and Architecture}

One important characteristic of computational artifacts (such as robots) -- similar to biological agents -- is their dual nature: On the one hand their behavior (or function) can be observed from the outside, on the other hand this behavior is generated by a certain internal mechanism or algorithm \parencite{rahwan2019machine,schulte2018framework}.
The field of didactics of computer science differentiates therefore between the \emph{relevance} of a computational artifact -- which for the user is perceived as its function, e.g. the capability to autonomously drive in an environment or to execute spoken commands -- and the \emph{architecture} which is the algorithm or mechanism that produces this behavior e.g the processing chain from perception over reasoning to action making use of abstractions such as object categories and states. It has been postulated that it is important to make learners aware of the difference between relevance and architecture \parencite{schulte2018framework}. This is not an easy task as humans tend to have intuitive mechanisms how to predict other biological agents' behavior based on their own experiences.

\subsection{Relation to Human Interactive Learning and Pragmatic Frames}
\label{sec:pf}
When teaching children, humans exhibit a range of highly adaptive behaviors that tailor learning input to the learner's capabilities and understanding and facilitate learning by directing attention and structuring the interaction \parencite{brand2002evidence,nelson1989prosodic,vollmer2009motionese,pitsch2014tutoring}.
One important strategy applied by parents is the use of \emph{Pragmatic Frames}, recurring interaction patterns which allow the learner to use knowledge from known, previous interactions to new situations. This supports the process of abstracting from context \parencite{Bruner_PF, rohlfing2016alternative}. The potential of pragmatic frames for human-robot interaction has been described by \textcite{vollmer2016pragmatic}. When teaching a robot, humans seem to intuitively make use of pragmatic frames \parencite{hindemith2019pf}.
However, while using such strategies is beneficial for children, they may not be for current robot systems. For example, some learning algorithms rely on randomly sequenced learning data whereas teaching in context is based on repeating and slightly modifying an action again and again, thus leading to clusters of similar data in the input. Thus, these strategies are well tuned to the human's mind. This includes the ability for theory of mind (TOM), i.e. to take the perspective of another person, which requires a mental model of the other person consisting of her physical capabilities such as perception and action as well as her mental states such as goals and intentions and even emotional states \parencite{sterelny1990representational}. 
Humans tend to apply such a model to technical artifacts as well, as the involved processes are highly intuitive. However, as the cognitive architecture of technological artifacts such as robots is very different from a biological human mind, this often leads to misunderstandings and failed interactions. 
In this work, we therefore investigate in how far humans are able to benefit from technological concepts for the formation of correct mental models while interacting with robots.

\subsection{Communicating Technical Concepts}
An important assumption of this work is that communicating technical concepts of the robot helps shaping the mental model users have about robots and improves the understanding of the human-robot interaction. Communicating information for human-robot interaction is usually done via instructions that should be as intuitive and accessible to non-experts as possible. Another strand of research communicates information directly via implicit or explicit robot feedback during the interaction.

\subsubsection{Instructions}
While experts know their system really well and are able to interact with it, without additional information, naive users fail to do so. Therefore it is necessary to provide new users with supplementary materials to help them to understand the system better. These additional information can be provided in various ways. For example, information can be given in the form of a manual to read, a video to watch, or a tutorial where users take also actions. \textcite{cakmak2014teaching} investigated the influence of these three channels on the users' ability to successful interact with a robot. The different types of instructional material were tested in a \emph{Programming by Demonstration} scenario. The results show that interaction videos mediate the needed knowledge the best. Based on these results, we also utilize interaction videos in our study.

\subsubsection{Feedback}
For the literature on robot feedback for robot learning, one body of work is concerned with robot feedback about the course of interaction and the robot's current state.
\textcite{BERT} showed that a robotic system that gives feedback about its current state but fails at some point, is preferred over a system that does not give feedback but works reliably.
\textcite{RT} developed a graphical representation of the decision and action making process of the robot for transparency and showed that it helped users to understand the decision making process of the robot.

Other work on feedback focuses on the perception capabilities of the robot.
\Textcite{Breazeal} showed that a combination of explicit and implicit feedback improves the effectiveness of human-robot interaction because malfunction of the robot can be more easily detected and recovered, in contrast to only explicit feedback.
\Textcite{Thomaz09} tested the learning performance of a robot that learned objects in a social way and a non-social way. In the social condition, the robot used gaze behavior to indicate errors, which besides better sampling from human partners led to faster error recovery in the interaction.
While this paper focused on implicit feedback, \textcite{Otero08} examined the impact of explicit verbal feedback on the perception of user demonstrations. The majority of the subjects repeated the same demonstration until positive feedback was given by the robot, instead of trying to optimize the given demonstration.

Other approaches on robot feedback aim to communicate what the robot learned (e.g., \cite{vollmer2014robots,Greef15}) and its execution capabilities (e.g., \cite{ERI}).

\section{Concepts and Hypotheses}
\label{sec:CaH}

Based on the goal of improving human-robot interactions by shaping an appropriate mental model of the robot, we hypothesize the following:

\paragraph{Hypothesis 1:} Providing architectural concepts allows users to gain more knowledge about the functionality of a robot.

\paragraph{Hypothesis 2:} Insights into the architecture of a robot increases the ability to recognize and explain errors in human-robot interaction.

\paragraph{Hypothesis 3:} Technical concepts differ in terms of their familiarity and observability. These factors influence the user's ability to recognize and understand problems in human-robot interactions.

\section{Methods}
\label{sec:meth}
In order to test our hypotheses, we developed two ways of providing architecture insights about the robot: a) architecture information is given in an \emph{instruction video} prior to the human-robot interaction and b) a \emph{visualization} of the current internal states of the robot is shown to users along the interaction with the robot. 

We conducted an online study devised as a survey. In this survey, participants watched different erroneous human-robot interaction videos and answered questions regarding the source of the underlying problems. We argue that the ability to detect and explain errors in these interactions implies an improved user's mental model about the robot. The participants were randomly assigned to one of four conditions in which they were provided with different amounts and types of information:
\begin{itemize}
    \item \texttt{Rel}: the baseline (without architecture information and thus, only with relevance information)
    \item \texttt{Arch}: architecture (architecture information communicated via an instruction video before the interaction)
    \item \texttt{Vis}: visualization (architecture information communicated via an online visualization during the interaction)
    \item \texttt{Arch+Vis}: architecture \& visualization (architecture information communicated both, via an instruction video before the interaction and an online visualization during the interaction)
\end{itemize}

\subsection{Scenario}
\label{sec:scen}
For investigating our hypotheses, we developed our system with regard to an object learning scenario. The goal of this scenario was to teach the robot a label for an object. The object recognition was simplified by using Aruco marker detection \parencite{ARUCO}. These markers were attached to the objects. In that way, each object could be uniquely identified. The learning of a label for an object was realized by storing a map between an Aruco marker ID and the verbally provided label. We decided on this scenario because we use concepts that are common in human-robot interactions and are important for users to understand. In the following, we will give you a more in-depth view on the used system and the mentioned concepts.

\subsection{System and Concepts}
Our approach was realized on a robotic system. Based on the setup of the robot and the used scenario, we decided on the technical concepts that should be communicated.

\subsubsection{Robot}

\paragraph{Robotic Platform}
We used the robotic platform \emph{scitos G5} by Metralabs \parencite{metralabs}. A model of the robot can be seen in \cref{fig:archVideo}. The robot is equipped with two RGB-D cameras. One mounted on a pan-tilt unit at the top of the robot. The other one below the tablet. To display information, a touch display is mounted in front of the robot. Behind this display a microphone and two speakers are positioned. The robot is also equipped with a 6 DOF robotic arm and a mobile base with front and rear lasers, which were not used in this study.

\paragraph{System}
\label{par:System}
The robot was developed based on ROS \parencite{ros}. The robot's main functionalities for the scenario were:
\vspace{-5mm}
\begin{itemize}
    \itemsep0em
    \item object recognition
    \item speech recognition
    \item speech synthesis
    \item behavior control
\end{itemize}

The \emph{object recognition} was developed using the Aruco marker detection to overcome the problem of unstable object recognition. The RGB images from both cameras were used to detect these markers in the world. Objects were equipped with a marker, which had an unique identifier. Thereby the system was able to track each object, even if it was not visible for the robot for a certain amount of time. For the \emph{speech recognition}, we used the google service \emph{Cloud Speech-to-Text} \parencite{asr}. To reduce the amount of recorded audio by google, we used an additional wake-up-word detector. In order to interpret the recognized speech, we developed a grammar parser based on the Backus-Naur-Form \parencite{BNF}. The \emph{speech synthesis} was realized by a voice synthesizer. For the behavior control we developed a finite state machine, based on the flexbe engine \parencite{FLEXBE}.

\subsubsection{Concepts}
\label{sec:conc}

Based on the developed robotic system, we selected the three most common concepts in current robotic systems: object recognition, speech recognition and state machine.

\paragraph{Object Recognition}
For the concept of object recognition, we adopt a technical concept from a different subject area. 
This allows us to investigate the \textbf{third hypothesis}, which states that the familiarity and observability influences the ability to recognize and explain errors. Therefore, we did not use an object recognition system that detects the objects itself, but recognized objects by Aruco marker detection. The scanning of those markers is similar to scanning of QR codes \parencite{QR} or bar codes \parencite{BAR-Codes}.

\paragraph{Speech Recognition}
While our speech recognition software was able to recognize natural language, it was not able to infer the intend of the commands. Therefore we had to use a grammar parser to assign an intend to the speech command. Because the concept of having limited speech understanding capabilities is very important for users to know, we communicated this concept.

\paragraph{State Machine}
A commonly used way to control the robot's behavior is a finite state machine. In finite state machines, the robot's actions are realized as states. Each state has an outcome that can lead to other states. By designing such a set of states and connections between them, the robot is able to fulfill a predefined goal. The higher the flexibility of a robot's behavior should be, the more work it costs to design a corresponding state machine. Consequently, state machines are often limited in their flexibility and therefore the user has to stick to the pre-programmed sequence of actions in order for the robot to work properly. For the user to be able to follow the correct sequence, knowledge about the state machine and its sequence has to be provided.

Based on these three concepts we designed a human-robot interaction scenario that incorporates all of them. Furthermore, these concepts have been operationalized in error situations presented in videos (\cf~\cref{sec:scen}).

\subsection{Experimental design}
The design of communicating architecture information was based on the concepts mentioned above. For providing such insights about the robot, we decided on two complementary approaches. The first approach was to design an instruction video that gives direct explanations on the robot's architecture and functionality. This video is shown to users before the interaction with the robot.
The second approach communicates these concepts indirectly by providing visualizations about the internal states of the robot. This visualization is shown to users while they are interacting with the robot. In the following we will discuss each approach in more detail.

\subsubsection{Architecture Instruction Video} \label{subsec:AIV}
The instruction video was designed to communicate the technical concepts of the robot in a direct way. Accordingly, the video mediates the three concepts mentioned in \cref{sec:conc} without leaving room for interpretation. Our goal was to inform about the software and hardware features of the robot equally, while avoiding cognitive overload of the user. Hence, the video combined information about the hardware components and their functionality on the software level. To reduce possible confusions of the user, each concept was introduced independently from the other concepts. In order to direct the user's attention only to one aspect at a time, each fact about a concept was printed in a box and shown one after another. Previous boxes that correspond to the same concept were still displayed.
For an improved understanding of the used hardware, a 3-dimensional model of the robot was displayed in the center of the video. The corresponding hardware of each concept was also highlighted by a colored ellipses as an overlay of the 3-dimensional model. An arrow between the text box and the colored ellipses helped to connect the information. To improve the understanding of the displayed information, the text boxes are also read out. This results in stimulating the user in a multi-modal way and thereby improves the information reception \parencite{sweller2019cognitive}.

\begin{figure}
    \centering
    \includegraphics[width=\textwidth]{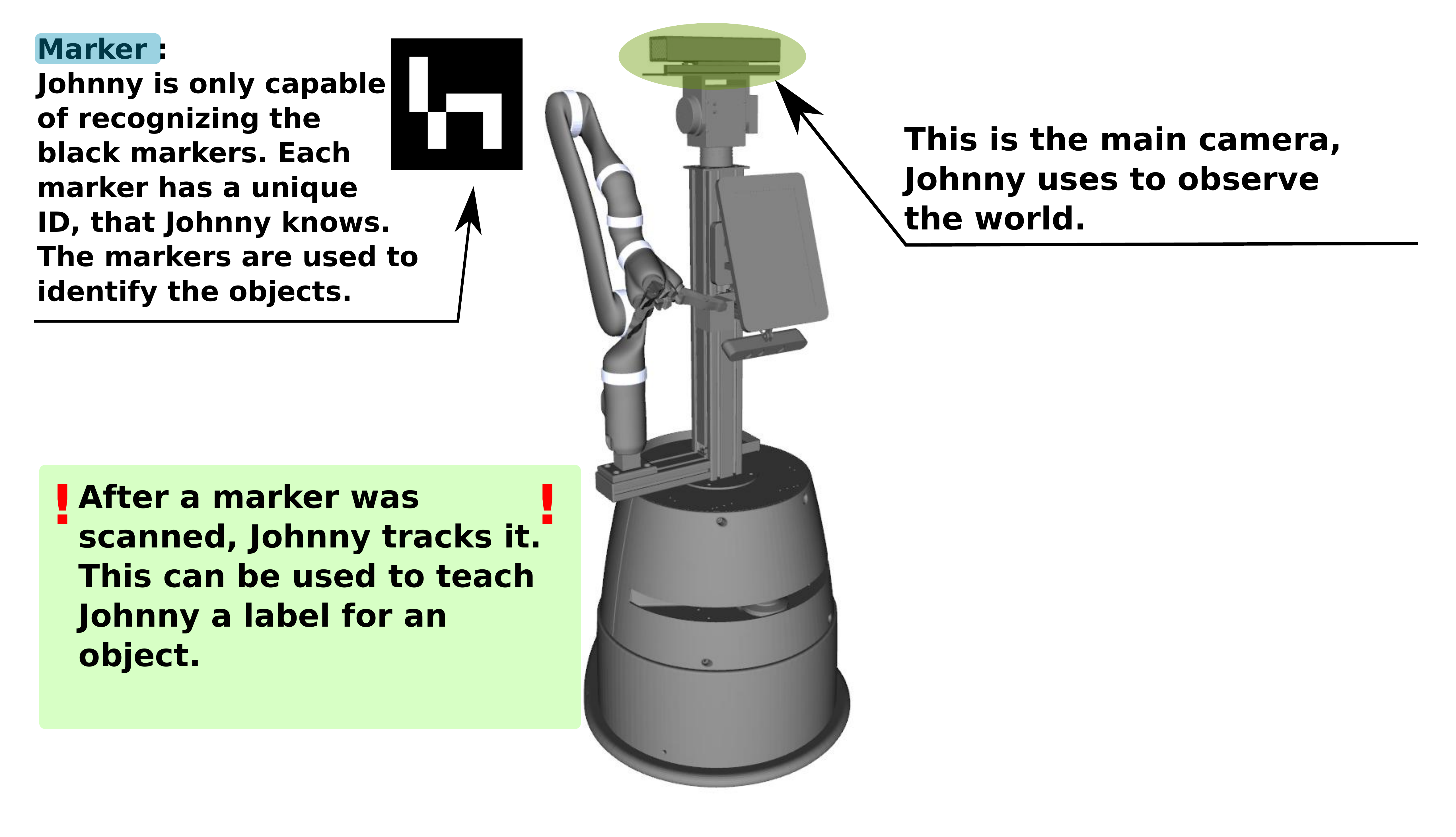}
    \caption{Sample scene from the instruction video that describes the camera of the robot. The text boxes were shown in the order: top right, top left, bottom left. The text in the top right describes the camera as a hardware feature. The text in the top left informs about the marker detection. How this marker detection can be used is described in the bottom left information box. }
    \label{fig:archVideo}
\end{figure}

The focus of the insights was to mediate the general concepts, their possibilities but also the limitations. No details about the underlying algorithms or their implementations were given. Hence, the information text was designed to contain technical terms while still be simple enough to be understandable for naive users. In addition to this, the wording should not convey any analogies to human characteristics. For example, a wording such as \textit{\qq{With the microphone, the robot hears the user}} for the speech recognition module would imply that the robot can hear like a human. Instead, a wording such as \textit{\qq{The microphone [...] is used to process speech input from the user.}} emphasizes the technicality of the system. The text boxes of a concept were arranged to first mention the used hardware, followed by the software side usage. As a follow up, additional notes on how this information can be used was mentioned in a green highlighted box. Refer to \cref{fig:archVideo} for an example frame from the introduction video, in which the robot's camera is introduced together with object recognition.

{}

\subsubsection{Robot Visualization} \label{subsubsec:RS}
While the instruction video was shown prior to the interaction, the visualization of the internal states of the robot was shown alongside the interaction. In order to not disturb the interaction, the design communicated the technical concepts of the robot in a more indirect way. Consequently, this approach leaves room for interpretation on how the technical concepts work. As already described in \cref{sec:conc} we decided on three concepts that should be conveyed. In the following our visualizations of these concepts are described.

\begin{figure}
    \centering
    \includegraphics[width=0.3\textwidth]{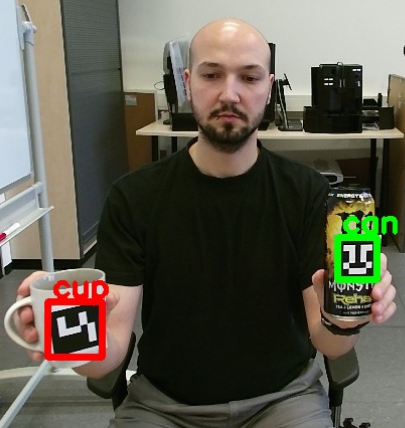}
    \caption{Clipped sample image of the object recognition visualization. The left marker is not tracked and therefore red. The right marker is currently tracked by the robot, which is indicated by a green bounding box. The learned label by the robot of each marker is written above the bounding boxes.}
    \label{fig:obj_rect}
\end{figure}

\paragraph{Marker Detection for Object Identification}
To communicate the visual perception of the robot, we showed the current image stream from the head camera of the robot. The processed information from the robot were visualized as an overlay of the image stream. The processed information is the Aruco markers detection. Therefore, the overlay are bounding boxes around the detected markers. The bounding boxes are colored depending on the current status of the marker. The status of a marker can be \emph{focused} or \emph{not focused}. A focused marker indicates that the current interaction is centered around this marker. If a marker is focused by the robot, the color of the associated bounding box is colored green. Otherwise, the bounding box is colored red. The design choice of the colors are the same as for the speech recognition (\cf~\cref{par:vercom}). If a label for an Aruco marker was stored in the memory, the label is printed above the corresponding bounding box and has the same color as the bounding box. Refer to \cref{fig:obj_rect} for an example.

\paragraph{Verbal Communication}
\label{par:vercom}
The main communication between the robot and the user was verbal. In order to trace the history of the interaction the visualization contained the utterances of both the robot and the user. Because the process of speech synthesis by the robot had no influence on how the user is expected to behave, no further information were provided about this concept.

In contrast to the speech synthesis, to better understand the process of speech recognition, a visualization has to communicate three pieces of information. Because the robot is not constantly listening, it is important for a user to be able to make out and understand whether the robot is listening to speech input at any given point in time. After an utterance was recognized, a visualization should show \emph{what} was recognized and \emph{whether} the command could be interpreted by the robot. A visualization that communicates these pieces of information enables the user to detect potential errors in the speech recognition process.

\begin{figure}
    \centering
    \includegraphics[width=0.75\textwidth]{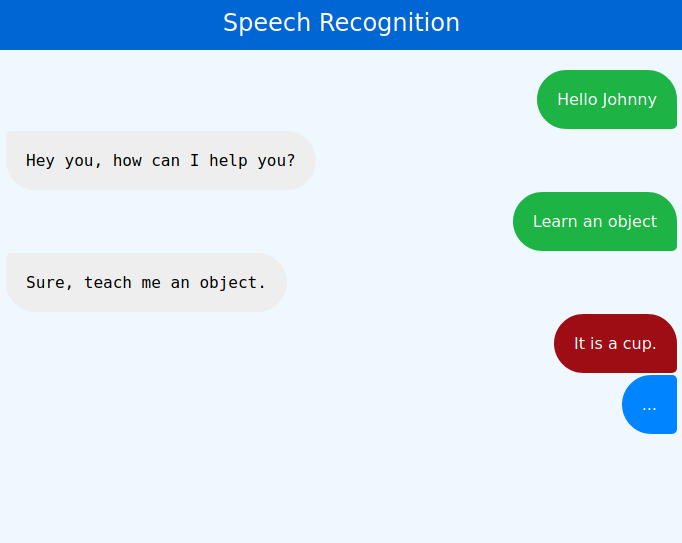}
    \caption{The chat box visualization for speech recognition and speech synthesis. On the left is written what the robot said. The right side displays the speech recognition of the robot. If the robot currently listens to speech is indicated by a blue speech bubble with three dots. The recognized speech is written in a speech bubble on the right side as well. If the speech could be parsed by the robot, it is displayed in a green speech bubble (e.g. first and second recognized speech). Otherwise it is written in a red speech bubble (e.g. last recognized speech).}
    \label{fig:chatbox}
\end{figure}

The overall layout was designed in an intuitive and familiar way. Accordingly we decided to visualize the verbal communication in the style of a chat box, such as WhatsApp \parencite{whatsapp} or Messenger \parencite{messenger}. With $2 \times 10^9$ monthly users for WhatsApp and $1.3 \times 10^9$ monthly users for Messenger in October 2019 \parencite{messengerUse} the design of chat boxes are familiar to a large percentage of the population. Our design, with exemplary input can be seen in \cref{fig:chatbox}. The arrangement of the text boxes was based on the perspective of the user. Therefore, the speech output of the robot was printed in a speech bubble on the left side of the chat box. The right side shows the speech recognition of the command provided by the user. This allocation of perspective is the de facto standard in messengers with such a design.
While the robot is listening to speech input from the user, a blue speech bubble with three dots is displayed. This visualization is used by the WhatsApp messenger to communicate that someone is currently writing a message. We used this to symbolize the listening state of the robot. After the speech input is processed by the robot, the resulting recognition is displayed. At this point we differentiate between speech that was recognized and where an intend could be determined and those which could not be interpreted. If the robot could determine an intend, the speech bubble of the corresponding speech input had a green background color. Otherwise the background color of the speech bubble was red. We decided on these colors, because in our modern society green and red are often occupied with \emph{positive} respectively \emph{negative} meaning. For example, traffic lights use the color green to communicate that a driver is allowed to drive, respectively red to communicate the driver must wait. In addition, the color red mediates possible danger.

\begin{figure}
    \centering
    \includegraphics[width=\textwidth]{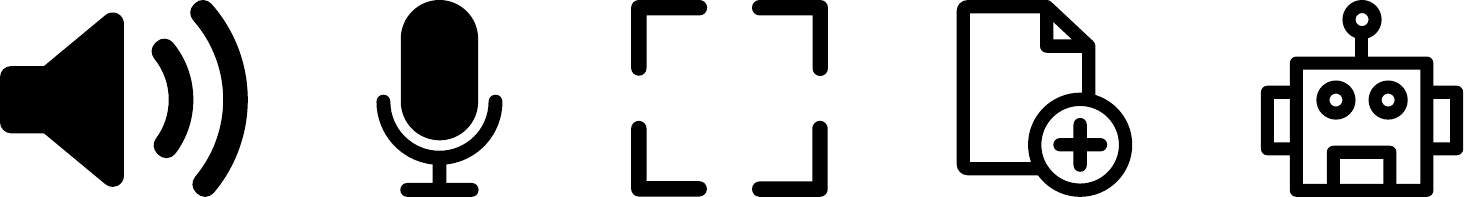}
    \caption{The used icons to visualize the current state of the robot. The icons from left to right: The robot is currently speaking (\texttt{robot speaks}), the robot listens for speech input (\texttt{robot listens}), the robot scans the room for a marker (\texttt{robot scans}), the robot stores new information in its memory (\texttt{robot stores}), the robot is currently occupied with a task (\texttt{robot works})}
    \label{fig:icons}
\end{figure}

\paragraph{Finite State Machines for Robot Control}
Based on the highly mathematical nature of finite state machines, we expect this technical concept to be incomprehensible of most naive users. Therefore, our goal was to visualize a simplification of this process. Instead of visualizing the entire sequence, we only visualized the current state. In order to make the current state comprehensible, while requiring as little attention as possible, each state was represented as an icon. \cref{fig:icons} show the five icons used in our visualization to communicate the most important states of the state machine.

The designs for the icons were made by \emph{kiranshastry} from \emph{www.flaticon.com}. Our approach was to use icons that are simple, follow a uniform design among each other and are intuitive. Based on these guidelines, the symbols for speech recognition (\texttt{robot listens}) and robot is speaking (\texttt{robot speaks}) are based on symbols used by Microsoft's operating system Windows, which is the most popular operating system worldwide \parencite{OS}. Since QR markers are used in many areas of application, the scanning symbol (\texttt{robot scans}) is familiar to many people. The symbol to store new information (\texttt{robot stores}) is a combination of a document and an \emph{add} (plus) symbol. This combination of two familiar icons facilitates an intuitive understanding. The last symbol is a robot and is used whenever the robot is doing something where the user needs to wait (\texttt{robot works}). While this design might not be known from somewhere else, a robot face can be easily identified while still be very simple in its design.

\subsubsection{Course of Online Study}
To verify our hypotheses we conducted an online survey with 123 participants. In this survey, participants had to detect and further elaborate on erroneous human-robot interactions in videos. Furthermore, parts of the participants' mental models about the robot were measured. We divided the participants in four different condition groups. Each group was shown different amount of information about the robot. To measure the impact of providing technical concepts of the robot to the participants, we used a $2\times2$ study design. As a baseline, the \texttt{Rel} group were only shown the interaction video with no additional information. The \texttt{Arch} group received additional information about the robot's features as an instruction video at the beginning. In an indirect way, the group \texttt{Vis} was shown a visualization of the internal states of the robot along the interaction video. As a fourth group, \texttt{Arch+Vis} were shown the instruction video together with the enriched interaction videos. The composition of the condition groups can be seen in \cref{tab:conditions}.

\begin{table}
    \centering
    \rowcolors{2}{gray!15}{white}
    \caption{The condition groups of the online survey. The \texttt{Rel} condition were shown neither an instruction video nor a visualization. The \texttt{Arch} condition were shown an instruction video. The \texttt{Vis} condition received a visualization of the robot's internal states along the interaction video. The fourth condition \texttt{Arch+Vis} were shown an instruction video as well as a visualization.}
    \begin{tabular}{l c c}
        \toprule
        \rowcolor{gray!5}
        Condition Type & No Instruction & Instruction  \\
        \midrule
        No Visualization & \texttt{Rel} & \texttt{Arch} \\
        Visualization & \texttt{Vis} & \texttt{Arch+Vis} \\
        \bottomrule
    \end{tabular}
    \label{tab:conditions}
\end{table}

\begin{figure}
    \centering
    \includegraphics[width=1\textwidth]{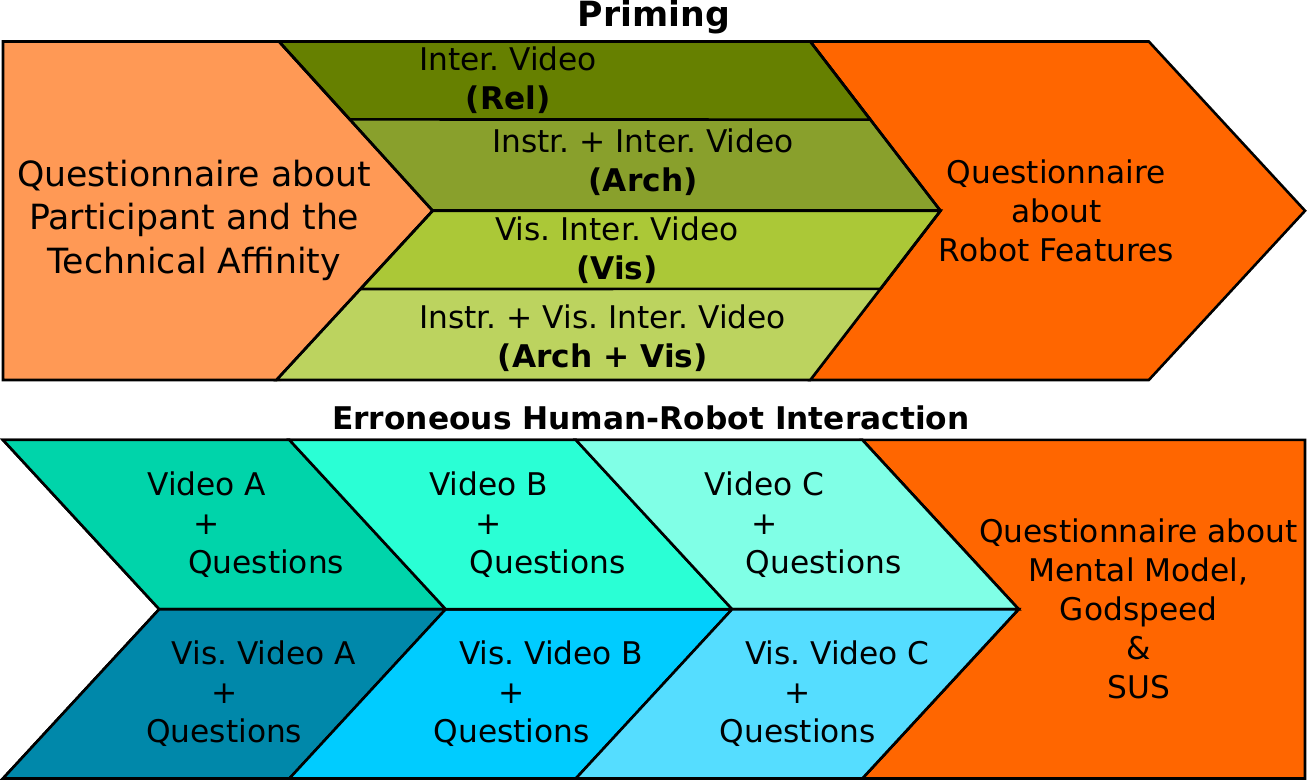}
    \caption{Course of online study, starting from top left to bottom right. At the beginning, participants were asked to fill out questionnaires about themselves and their technical affinity. Afterwards, depending on the condition, the participants were shown an instruction video and an interaction video. In a follow up questionnaire about the robot features, the participants knowledge about the robot was surveyed. Afterwards, the erroneous interaction videos were shown and related questions were asked. In the end, questionnaires about the mental model, \emph{godspeed} and \emph{SUS} were surveyed.}
    \label{fig:study_timeline}
\end{figure}

\begin{figure}
    \centering
    \includegraphics[width=0.7\textwidth]{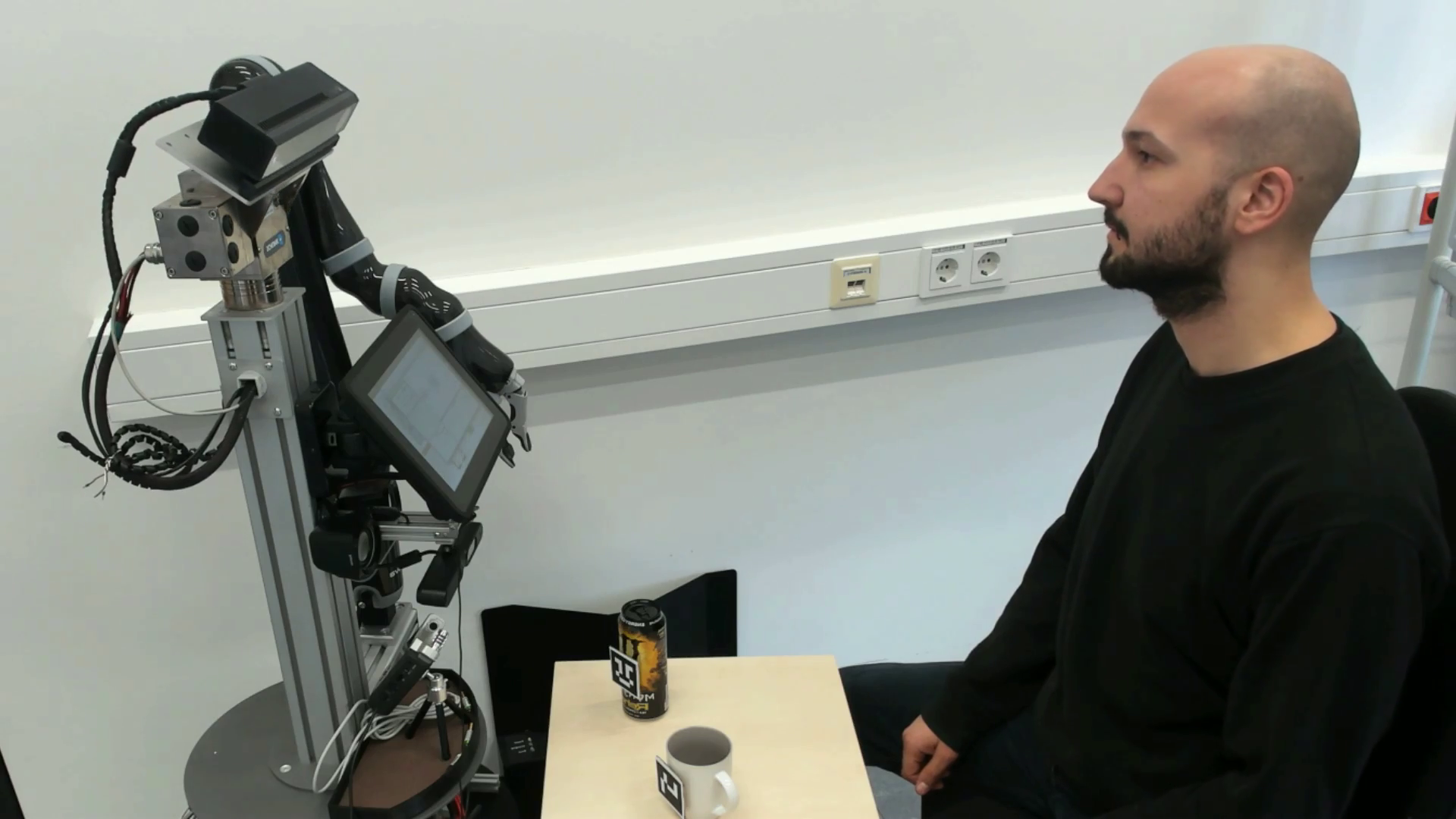}
    \caption{Frame of the interaction video.}
    \label{fig:normalVid}
\end{figure}

\begin{figure}
    \centering
    \includegraphics[width=0.7\textwidth]{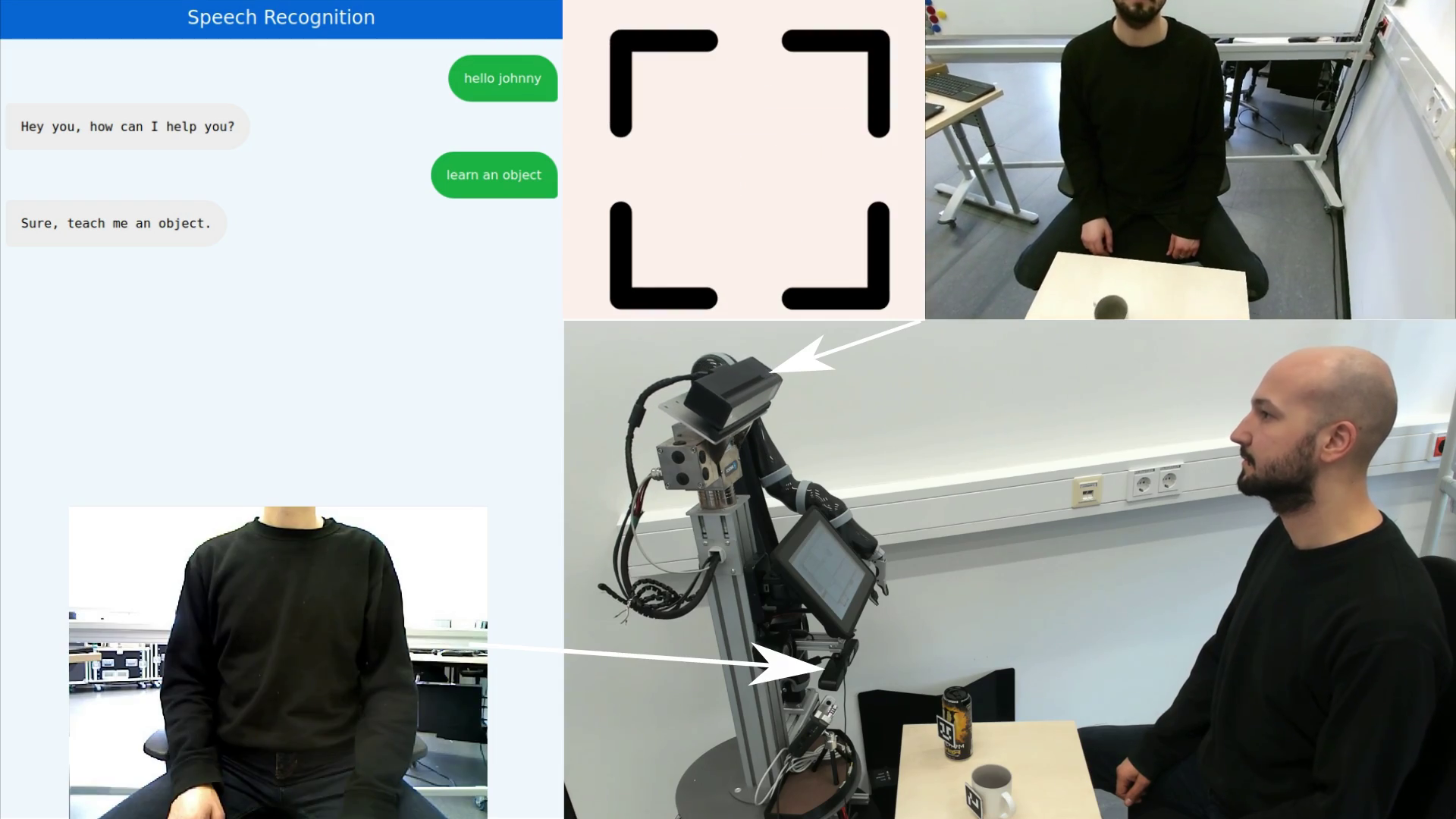}
    \caption{Frame of the interaction video with additional visualizations, describing the internal states of the robot. The speech recognition visualization was displayed in the top left. The robots camera stream was displayed in the top right and bottom left, respectively. The current state machine state was displayed in the top center. The actual human-robot interaction was displayed in the bottom right.}
    \label{fig:visVid}
\end{figure}

The online survey was carried out using the online portal \emph{Prolific} \parencite{prolific} to acquire participants. The course of the survey is illustrated in \cref{fig:study_timeline}. First, the participants were welcomed and asked general questions about their person. Furthermore, we measured their technical affinity, using the \emph{Affinity for Technology Interaction Scale} \parencite{franke2019personal}. After the general questionnaires, each participant was randomly assigned to one of the four condition groups. Depending on the condition to which the participants were assigned, they were shown an instructional video with architectural information about the robot or not (\cf~\cref{subsec:AIV}). Afterwards a video of a successful human-robot interaction, following the scenario, was presented. Depending on the condition, the interaction video was enriched with information about the internal states of the robot (\cf~\cref{subsubsec:RS}). In order to check the participants knowledge about the robot's features, open questions about features of the robot were asked. We separated features of the robot into \emph{hardware} and \emph{software}. Afterwards, three erroneous human-robot interaction videos were presented to the participants in a randomized order. While the interaction in the videos stayed the same for all conditions, the \texttt{Vis} and \texttt{Arch+Vis} conditions were shown videos that were enriched in the same way as the initial interaction video. We refer to \crefrange{fig:normalVid}{fig:visVid} for exemplary frames from one of the erroneous interactions. After each video, several questions about the origin of the error and how participants observed it were asked.
Subsequently, each participant was asked open questions about the technical concepts that led to the errors and what they associate with them. In addition, parts of the \emph{Godspeed} questionnaire \parencite{GOD} and the \emph{SUS} questionnaire \parencite{SUS} were collected.\\

\subsubsection{Human-Robot Interaction Videos}
\label{sec:interaction}
\begin{figure}
    \centering
    \includegraphics[width=\textwidth]{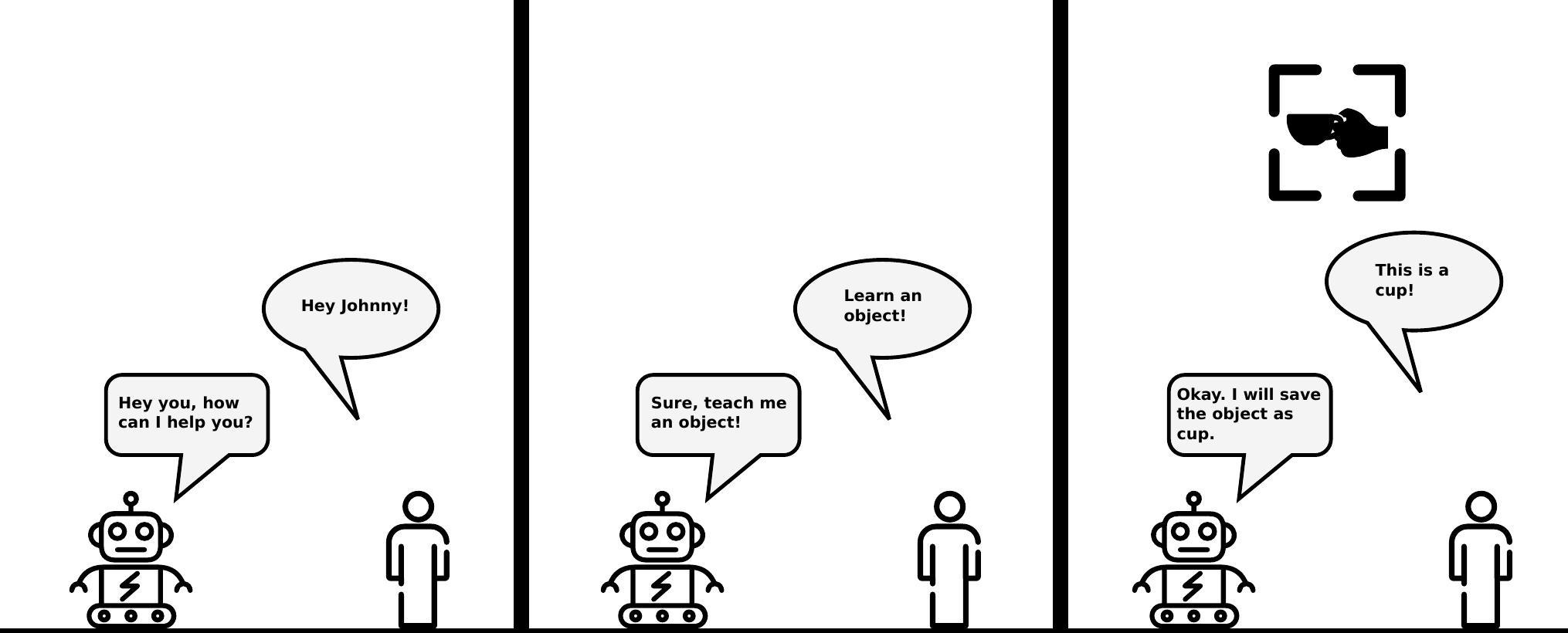}
    \caption{The interaction flow of the object labeling scenario, from left to right. First, the human greets the robot, whereupon the robot welcomes the human and asks for a task. The human triggers the object learning interaction by a command, which the robot confirms. To teach a label for an object, first the human shows the object of interest into the scanning camera of the robot. Afterwards the human provides the label of the object verbally. The robot confirms the learning verbally and ends the interaction.}
    \label{fig:interactionFlow}
\end{figure}

For the human-robot interaction, we have implemented an object labeling scenario (\cf~\cref{sec:scen}). The general interaction flow is shown in \cref{fig:interactionFlow}. In a first step, the user starts the interaction by greeting the robot. The robot welcomes the user and asks for a task. The user then activates the learning phase by giving the verbal command to learn an object by saying \emph{\qq{Learn an object!}}. The robot confirms the command by saying \emph{\qq{Sure, teach me an object!}}. subsequently, the pan-tilt unit tilts down, so that the table between the human and the robot can be captured by the camera. The robot then waits for an object to be visible in its bottom camera. First, the human points a cup into the camera of the robot while facing the marker towards the camera. Afterwards, the label is verbally provided by saying \emph{\qq{This is cup!}}. The robot confirms the learning by saying \emph{\qq{Okay, I will save the object as cup.}} and stores the information in its memory. For the erroneous interaction videos, this interaction flow is violated at some point each. But in each video there is no more than one error.

To test the hypotheses, we asked the participants whether they were able to detect the error in the interaction of the video. For this, the same human-robot interaction as described in \cref{sec:scen} was repeated, but a mistake was induced which led to an erroneous interaction. The errors were based on the concepts described in \cref{sec:conc}. The erroneous interaction videos were divided into three videos, where each video only contained one error. All mistakes were induced by the human who was not aware of the limitations of each concept.\\

\paragraph{Object Detection Error}
One of our technical concepts was that the robot only can detect Aruco markers to identify objects. To generate an error that aims at this concept, the user fails at pointing the marker towards the robot and instead rotates the marker sideways in front of the external camera. Because the robot cannot detect any objects it encounters a timeout while waiting for an object. This in turn causes the robot to switch to a \qq{failed} mode, where it says \emph{\qq{Sorry, but something went wrong!}} then ends the interaction. In the following we refer to this video with \texttt{No Object}.

\paragraph{Speech Recognition Error}
Because the robot can only process exact commands, we induced an error by rephrasing the utterance to provide the label. Therefore, instead of saying \emph{\qq{This is a cup}}, the user says \emph{\qq{It is a cup}}. Both sentences have the same meaning while the wording differs. Because the robot cannot process the command, it switches to a \qq{failed} mode and says that something went wrong, in the same way as for the object detection error, then ends the interaction. In the following we refer to this video with \texttt{Failed Speech}.

\paragraph{State Machine Error}
For the concept of state machines, the interaction sequence is violated at some point. After the robot switches to the \qq{learning} mode, the robot expects to scan an object. Instead, the user first provides the label and tries to show the object afterwards. The robot switches to a \qq{failed} mode after receiving the label, communicates that it encountered a problem and ends the interaction. In the following we refer to this video with \texttt{Failed SM}.

\section{Results}
\label{sec:res}
The analyses were carried out on basis of 122 out of 130 participants. 8 participants had to be excluded due to incomplete entries. All participants were located in the United States (63 women, 56 men, 2 diverse, 1 anonymous, $M_{age}=34.09$ years, age range: 18-74). Each participant was assigned randomly to one of the four condition groups (24 \texttt{Rel}, 38 \texttt{Arch}, 29 \texttt{Vis} and 31 \texttt{Arch+Vis}). Due to the randomization process of the survey software, condition \texttt{Rel} has slightly less and condition \texttt{Arch} has slightly more participants than the other groups. One-way ANOVA tests \parencite{ANOVA} for gender and age distribution did not indicate any significant differences between the condition groups. Hence, the condition groups are balanced in terms of their participants.

\subsection{Hypothesis 1: \emph{Providing  architectural  concepts  allows  users  to  gain  more  knowledge  about  the functionality of a robot.}}

To investigate this hypothesis we asked all participants questions regarding the \emph{hardware} and \emph{software} features of the robot after they have seen the introduction. To analyze the answers to the open questions we assigned each one of them the features mentioned. We only focused on features that were important for the human-robot interaction in the videos. As key figures we calculated the number of features each participant mentioned. In a second step we took a detailed look at each feature and how many participants mentioned them.

\paragraph{\emph{Which components does the robot have that allow it to observe or interact with its environment?} (hardware)}

The important features for the hardware of the robot were:
\vspace{-5mm}
\begin{itemize}
    \setlength\itemsep{2mm}
    \item camera
    \item microphone
    \item pan-tilt unit (PTU)
    \item speaker
\end{itemize}
In order to observe the influence of the instruction video on the knowledge about the robot, the \emph{speaker} was not mentioned in the introduction video but was included in the analyses.

We first applied a one-way ANOVA, which indicated a significant difference in the number of mentioned hardware features. A follow up post-hoc tukey-HSD test \parencite{tukeyHSD} revealed that the condition \texttt{Arch+Vis} mentioned significantly more than the \texttt{Rel} condition (p-value = 0.0214) and had a tendency of increase for the \texttt{Vis} condition (p-value = 0.0807) (\cf~\cref{tab:num_hw}).\\

\noindent
\begin{minipage}{0.5\textwidth}
    \centering
    \includegraphics[width=\textwidth]{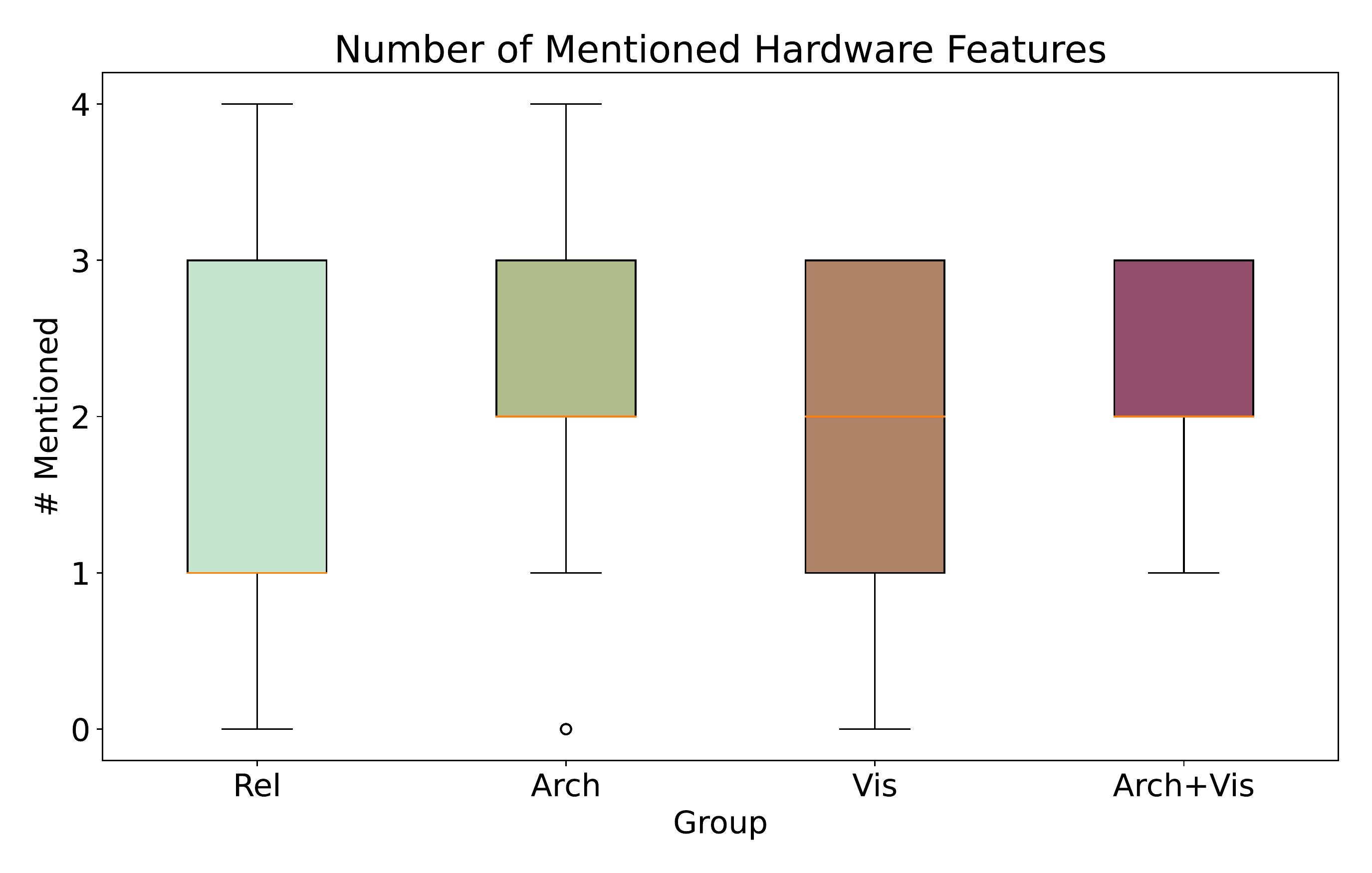}
    \captionof{figure}{Number of mentioned hardware features per participant.}
    \label{fig:hw_num_mentioned}
\end{minipage}%
\vspace{1em}
\begin{minipage}{0.5\textwidth}
    \centering
    \vspace{-2em}
    \includegraphics[width=\textwidth]{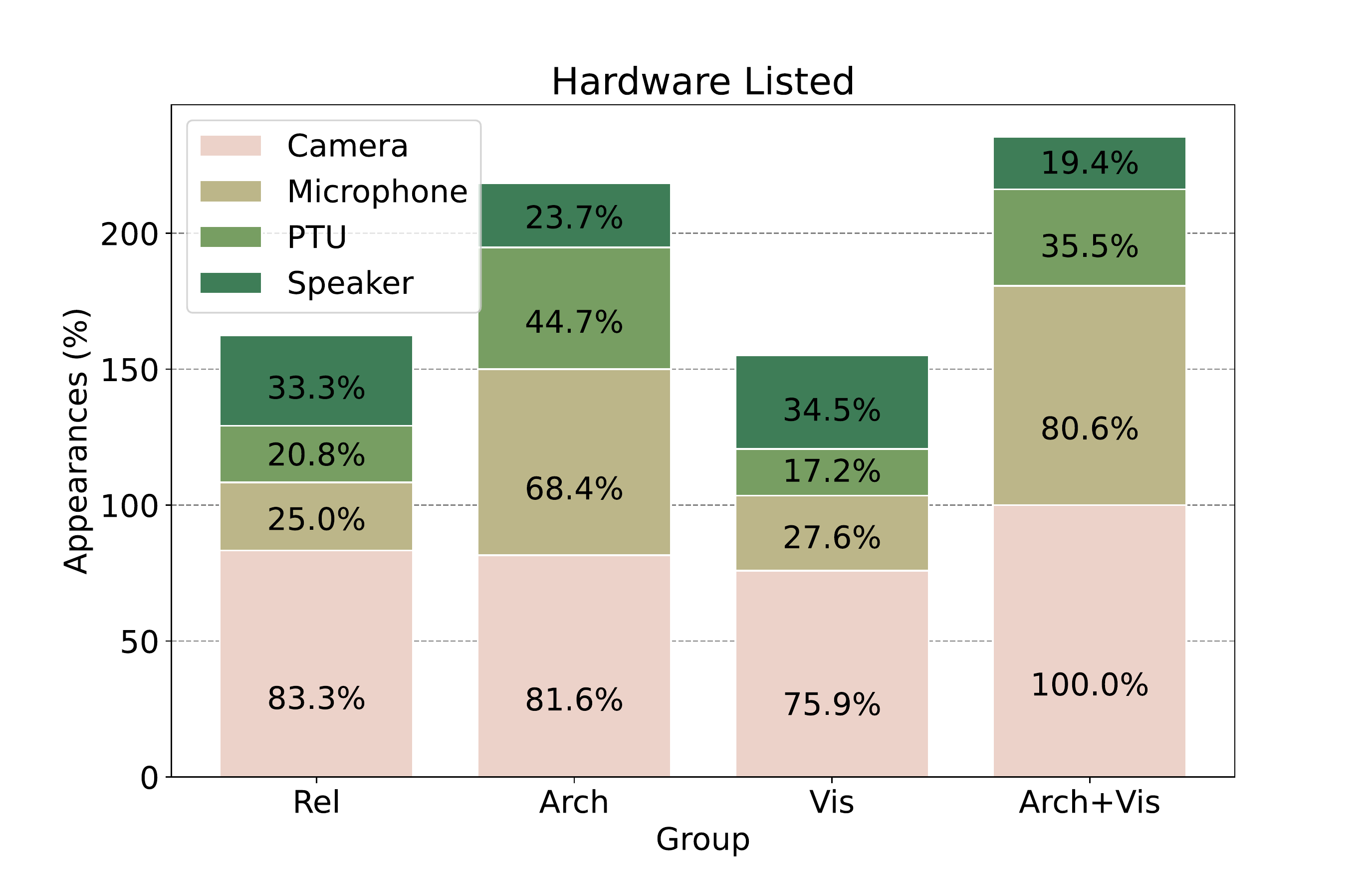}
    \captionof{figure}{Frequency of each hardware feature.}
    \label{fig:hw_stats_perc}
\end{minipage}%

\begin{table}
    \centering
    \caption{One-Way ANOVA, with follow up post-hoc tukey-HSD test for the number of hardware features mentioned between all conditions. p-values $<$ 0.05 are highlighted in \textbf{bold}, p-values $<$ 0.1 are highlighted in \textit{italic}.}
    
    \begin{tabular}{l l S[table-format=1.4] S[table-format=1.4] S[table-format=1.4] S[table-format=1.4] c}
    \toprule
    \multicolumn{7}{c}{\textbf{One--Way ANOVA} (F-value = 4.1994, p-value = \textbf{0.0073})}\\
    \midrule
    Condition 1     & Condition 2 & {meandiff} & {p-adj} & {lower} & {upper} & reject \\
    \midrule
    \rowcolor{gray!15}
                                    & \texttt{Rel}      & 0.7298    & \textit{0.0617} & -0.0242   & 1.4838 & False \\
    \rowcolor{gray!15}
                                    & \texttt{Vis}      & 0.5592    & 0.1881 & -0.1639   & 1.2823 & False \\
    \rowcolor{gray!15}
    \multirow{-3}{*}{\texttt{Arch}} & \texttt{Arch+Vis} & -0.0733   & 0.9    & -0.8385   & 0.692  & False \\
    \multirow{2}{*}{\texttt{Vis}}   & \texttt{Rel}      & 0.1706    & 0.9    & -0.5005   & 0.8418 & False \\
                                    & \texttt{Arch+Vis} & -0.6325   & \textit{0.0807} & -1.3163   & 0.0513 & False \\
    \rowcolor{gray!15}
    \texttt{Arch+Vis}               & \texttt{Rel}      & 0.8031    & \textbf{0.0214} & 0.0867    & 1.5195 & True  \\
    \bottomrule
    \end{tabular}
    
    \label{tab:num_hw}
\end{table}

A closer look at each hardware feature (\cf~\cref{fig:hw_stats_perc}) turned out that the \emph{microphone} and \emph{PTU} features had differences in their frequency (Kruskal-Wallis test \parencite{Kruskal}). The follow up post-hoc Dunn test \parencite{dunn} revealed that the conditions \texttt{Arch} and \texttt{Arch+Vis} mentioned the \emph{microphone} feature significantly more frequently than the \texttt{Rel} and \texttt{Vis} conditions. Furthermore, the \texttt{Arch} condition mentioned the \emph{PTU} features more often than the \texttt{Rel} and \texttt{Vis} conditions (\cf~\cref{tab:hw_stats}).\\
The \emph{camera} and \emph{speaker} features had no differences in frequency between the conditions. In general, the \emph{camera} feature was mentioned by most participants, while the other features were mentioned fewer.

\begin{table}
    \centering
    \caption{Kruskal-Wallis test, with follow up post-hoc Dunn test for the frequency of mentioned hardware features. Below the features are the p-values of the Kruskal-Wallis test. Entries are p-values of the Dunn test. p-values $<$ 0.05 are highlighted in \textbf{bold}, p-values $<$ 0.1 are highlighted in \textit{italic}.}
    
    \begin{tabular}{l l S[table-format=1.4] S[table-format=1.4] S[table-format=1.4] S[table-format=1.4]}
    \toprule
    & & \multicolumn{4}{c}{{Hardware Feature}} \\
    \cmidrule{3-6}
    \multirow{2}{*}{Condition 1}    & \multirow{2}{*}{Condition 2} & {Camera}         & {Microphone}        & {PTU}               & {Speaker} \\
                                    &                              & {0.5317}& {\textbf{0.0000}}    & {\textit{0.0701}}   & {0.1971} \\
    \midrule
    \rowcolor{gray!15}
                                    & \texttt{Rel}                 & {-}              & \textbf{0.0001}            & \textbf{0.0303}            & {-} \\
    \rowcolor{gray!15}
                                    & \texttt{Vis}                 & {-}              & \textbf{0.0009}            & \textbf{0.0227}            & {-} \\
    \rowcolor{gray!15}
    \multirow{-3}{*}{\texttt{Arch}} & \texttt{Arch+Vis}            & {-}              & 0.9609            & 0.1462            & {-} \\
    \multirow{2}{*}{\texttt{Vis}}   & \texttt{Rel}                 & {-}              & 0.6017            & 0.9413            & {-} \\
                                    & \texttt{Arch+Vis}            & {-}              & \textbf{0.0012}            & 0.3423            & {-} \\
    \rowcolor{gray!15}
    \texttt{Arch+Vis}               & \texttt{Rel}                 & {-}              & \textbf{0.0002}            & 0.3913            & {-}  \\
    \bottomrule
    \end{tabular}
    \label{tab:hw_stats}
\end{table}

\paragraph{\emph{What skills and abilities does the robot have?} (software)}

As we did for the hardware features, we analyzed the participants answers regarding the appearance of the software features. Again, only the features important for the interaction videos were taken into account. These were:
\vspace{-6mm}
\begin{itemize}
    \setlength\itemsep{2mm}
    \item object recognition
    \item speech recognition
    \item speech synthesis
    \item learning
\end{itemize}

\cref{fig:sw_num_mentioned} illustrates how many software features each participant mentioned. A one-way ANOVA indicated no significant differences between the conditions (p-value = 0.0838).

\begin{minipage}{0.5\textwidth}
    \centering
    \includegraphics[width=\textwidth]{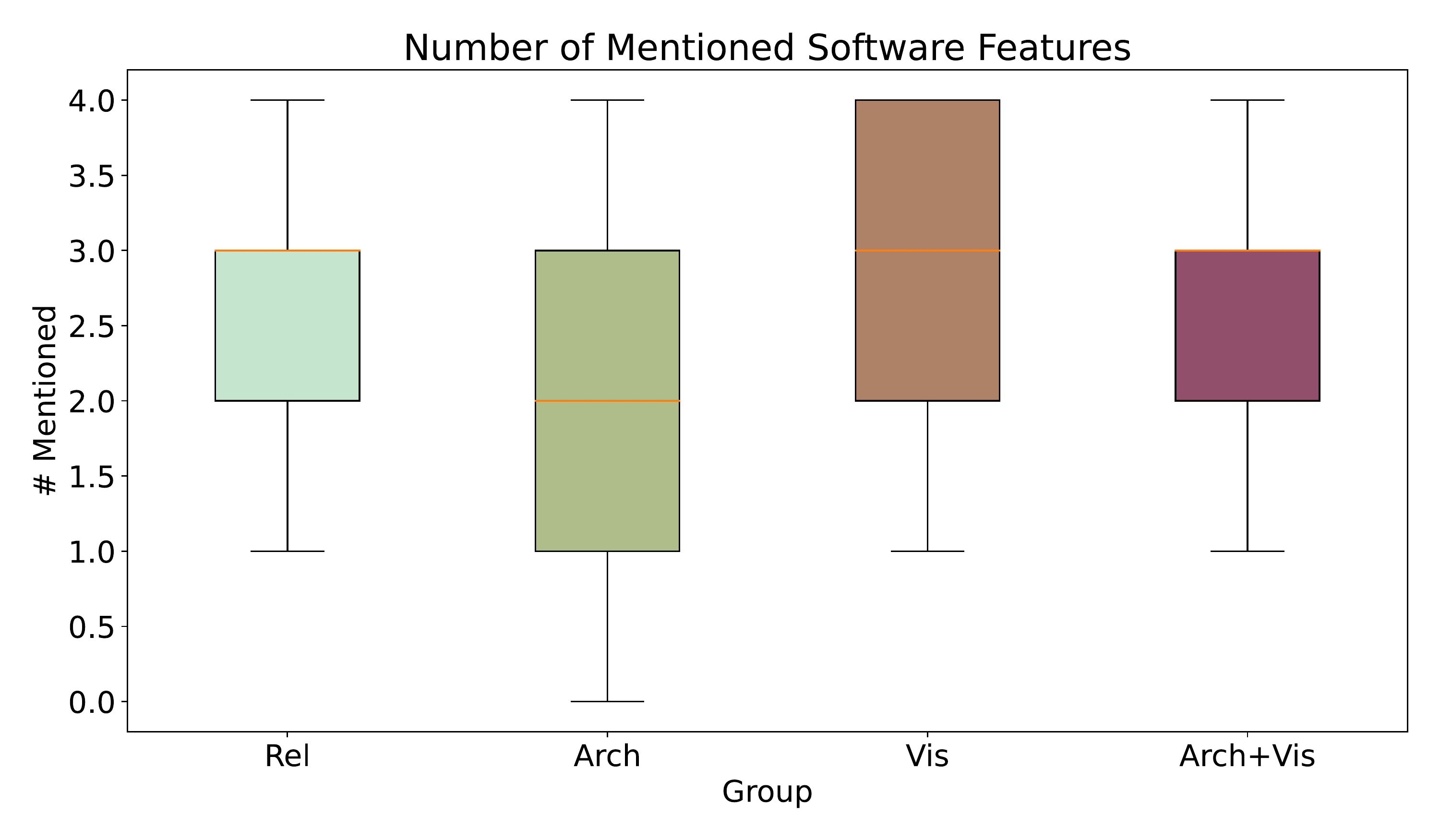}
    \captionof{figure}{Number of mentioned software features per participant}
    \label{fig:sw_num_mentioned}
\end{minipage}%
\vspace{1em}
\begin{minipage}{0.5\textwidth}
    \centering
    \vspace{-2em}
    \includegraphics[width=\textwidth]{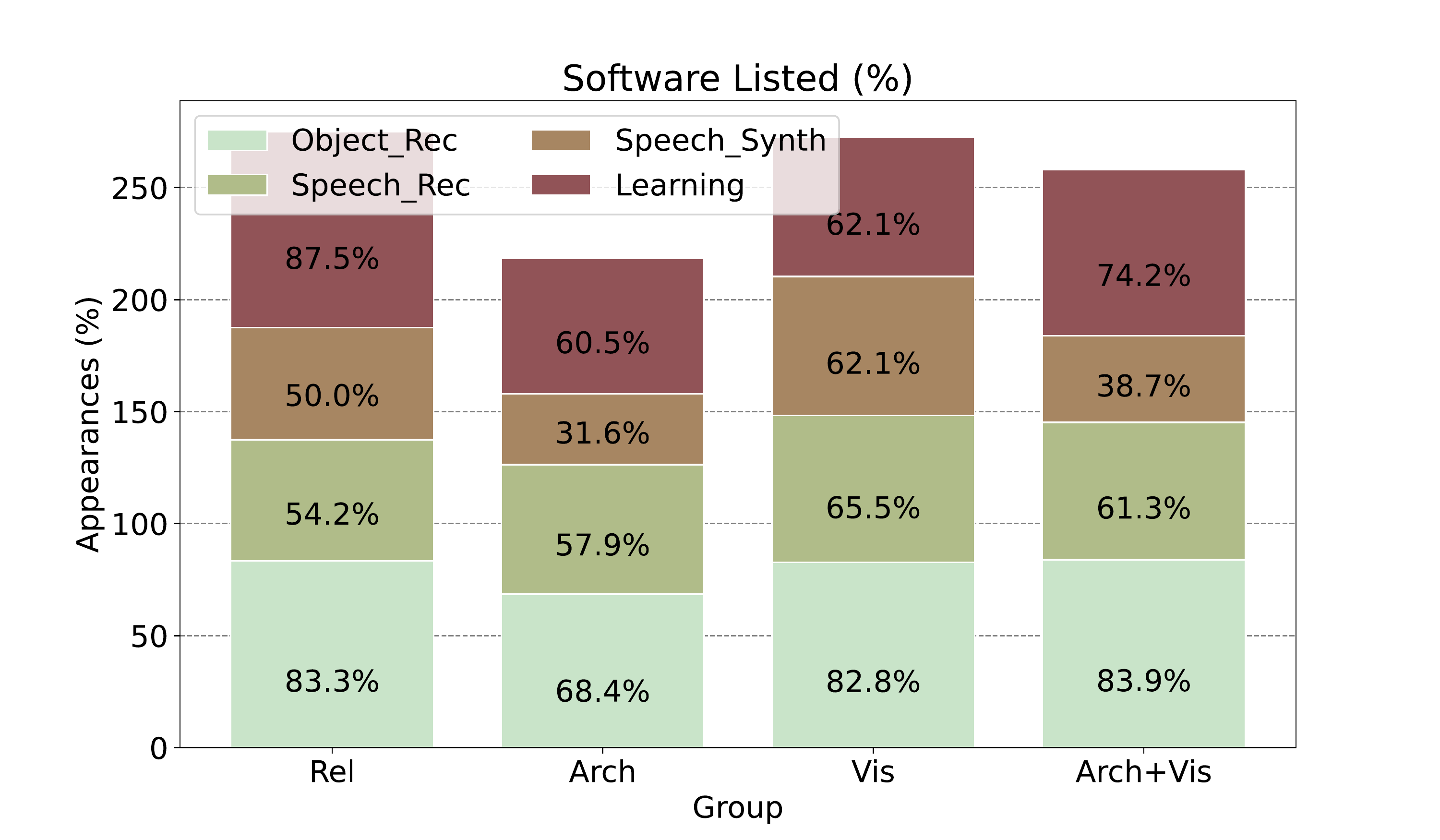}
    \captionof{figure}{Frequency of each software feature.}
    \label{fig:sw_stats_perc}
\end{minipage}%

We also analyzed how often each software feature was mentioned by the condition groups and applied the Kruskal-Wallis test with a subsequent Dunn test. As it can be seen in \cref{fig:sw_stats_perc}, all conditions mentioned the software features almost equally often. Tests showed only for the \emph{speech synthesis} that the \texttt{Vis} condition mentioned this feature significantly more frequently than the \texttt{Arch}  condition (p-value = 0.0167) and a trend in the difference for the \texttt{Arch+Vis} condition (p-value = 0.0701) (\cf~\cref{tab:sw_stats}).

\begin{table}
    \centering
    \caption{Kruskal-Wallis, with follow up Dunn test for the frequency of mentioned software features. Below the features are the p-values of the Kruskal-Wallis test. Entries are p-values of the Dunn test. p-values $<$ 0.05 are highlighted in \textbf{bold}, p-values $<$ 0.1 are highlighted in \textit{italic}.}
    
    \begin{tabular}{l l S[table-format=1.4] S[table-format=1.4] S[table-format=1.4] S[table-format=1.4]}
    \toprule
    & & \multicolumn{4}{c}{{Software Feature}} \\
    \cmidrule{3-6}
    \multirow{2}{*}{Condition 1}    & \multirow{2}{*}{Condition 2} & {Object Rec.}         & {Speech Rec.}        & {Speech Synthesis}               & {Learning} \\
                                    &                              & {0.4465} & {0.8674}    & {\textit{0.0912}}   & {0.1279} \\
    \midrule
    \rowcolor{gray!15}
                                    & \texttt{Rel}                 & {-}              & {-}            & 0.1793            & {-} \\
    \rowcolor{gray!15}
                                    & \texttt{Vis}                 & {-}              & {-}            & \textbf{0.0167}            & {-} \\
    \rowcolor{gray!15}
    \multirow{-3}{*}{\texttt{Arch}} & \texttt{Arch+Vis}            & {-}              & {-}            & 0.6055            & {-} \\
    \multirow{2}{*}{\texttt{Vis}}   & \texttt{Rel}                 & {-}              & {-}            & 0.3809            & {-} \\
                                    & \texttt{Arch+Vis}            & {-}              & {-}            & \textit{0.0701}            & {-} \\
    \rowcolor{gray!15}
    \texttt{Arch+Vis}               & \texttt{Rel}                 & {-}              & {-}            & 0.4055            & {-}  \\
    \bottomrule
    \end{tabular}
    \label{tab:sw_stats}
\end{table}

\subsection{Hypothesis 2: \emph{Insights into the architecture of a robot increases the ability to recognize and explain errors in human-robot interaction.}}

To examine this hypothesis, three erroneous human-robot interactions were shown to the participants. After each video, various questions regarding the error were asked. First we checked if the participants recognized the error and could explain why the error occurred. To analyze the answers to the open questions we chose a binary format to rate the correctness of each answer. The first analysis showed that if a participant could answer \emph{what} happened, the question regarding \emph{why} this error occurred was also answered correctly. Due to this, and the fact that some participants answered the \emph{what} question too general, we only focused on the explanations \emph{why} the error occurred, for further investigations.

\begin{figure}
    \centering
    \includegraphics[width=\textwidth]{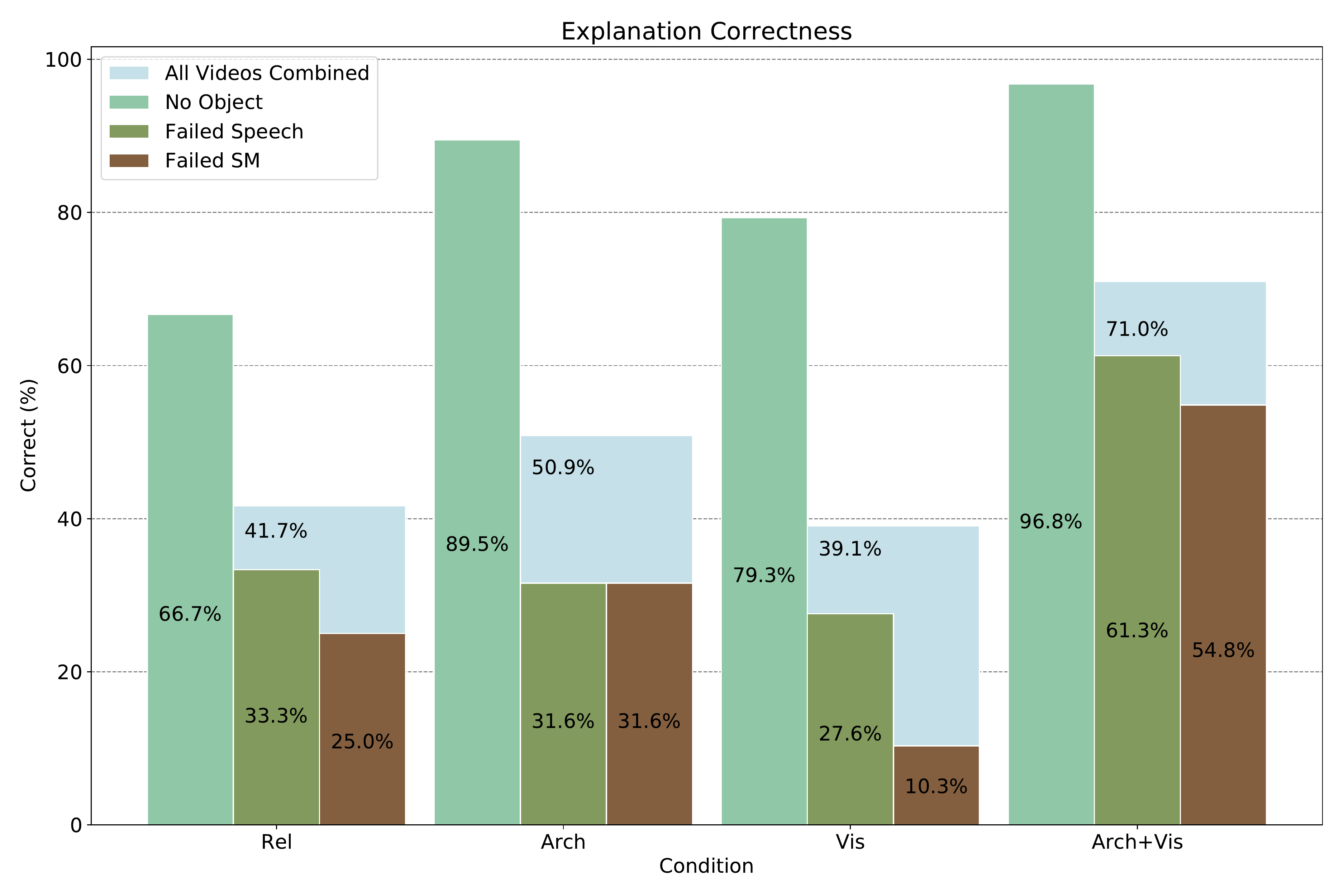}
    \caption{Correctness of the given explanations.}
    \label{fig:expl_correct}
\end{figure}

\cref{fig:expl_correct} shows how many participants provided correct explanations for each interaction video. The Kruskal-Wallis test indicated that for each erroneous video and also all videos combined, a significant difference between conditions could be found (\cf~\cref{tab:expl_correct}). While the post-hoc Dunn test already shows the most significant differences for the \texttt{Arch+Vis} conditions for each video, it surpasses all other conditions significantly in terms of number of correct explanations for the combined results of all videos.

\begin{table}
    \centering
    \caption{Kruskal-Wallis, with follow up Dunn test for the number of correct explanations. Below the videos are the p-values of the Kruskal-Wallis test. Entries are p-values of the Dunn test. p-values $<$ 0.05 are highlighted in \textbf{bold}, p-values $<$ 0.1 are highlighted in \textit{italic}.}

    \begin{tabular}{l l S[table-format=1.4] S[table-format=1.4] S[table-format=1.4] S[table-format=1.4]}
    \toprule
    & & \multicolumn{3}{c}{{Erroneous Interaction}}& \\
    \cmidrule{3-5}
    \multirow{2}{*}{Condition 1}    & \multirow{2}{*}{Condition 2} & {No Object}         & {Failed Speech}        & {Failed SM}               & {Combined} \\
                                    &                              & {\textbf{0.0142}} & {\textbf{0.0265}}    & {\textbf{0.0026}}   & {\textbf{0.0000}} \\
    \midrule
    \rowcolor{gray!15}
                                    & \texttt{Rel}                 & \textbf{0.0163}              & 0.8905            & 0.5874            & 0.2215 \\
    \rowcolor{gray!15}
                                    & \texttt{Vis}                 & 0.2576              & 0.7404            & \textit{0.0640}            & \textit{0.0978} \\
    \rowcolor{gray!15}
    \multirow{-3}{*}{\texttt{Arch}} & \texttt{Arch+Vis}            & 0.4074              & \textbf{0.0120}            & \textbf{ 0.0388}            & \textbf{0.0041} \\
    \multirow{2}{*}{\texttt{Vis}}   & \texttt{Rel}                 & 0.2082              & 0.6610            & 0.2534            & 0.7457 \\
                                    & \texttt{Arch+Vis}            & \textit{0.0634}              & \textbf{0.0076}            & \textbf{0.0002}            & \textbf{0.0000} \\
    \rowcolor{gray!15}
    \texttt{Arch+Vis}               & \texttt{Rel}                 & \textbf{0.0024}              & \textbf{0.0354}            & \textbf{0.0183}            & \textbf{0.0002}  \\
    \bottomrule
    \end{tabular}

    \label{tab:expl_correct}
\end{table}

\subsection{Hypothesis 3: \emph{Technical concepts differ in terms of their familiarity and observability.  These factors influence the user’s ability to recognize and understand problems in human-robot interactions.}}

\begin{figure}
    \centering
    \includegraphics[width=0.7\textwidth]{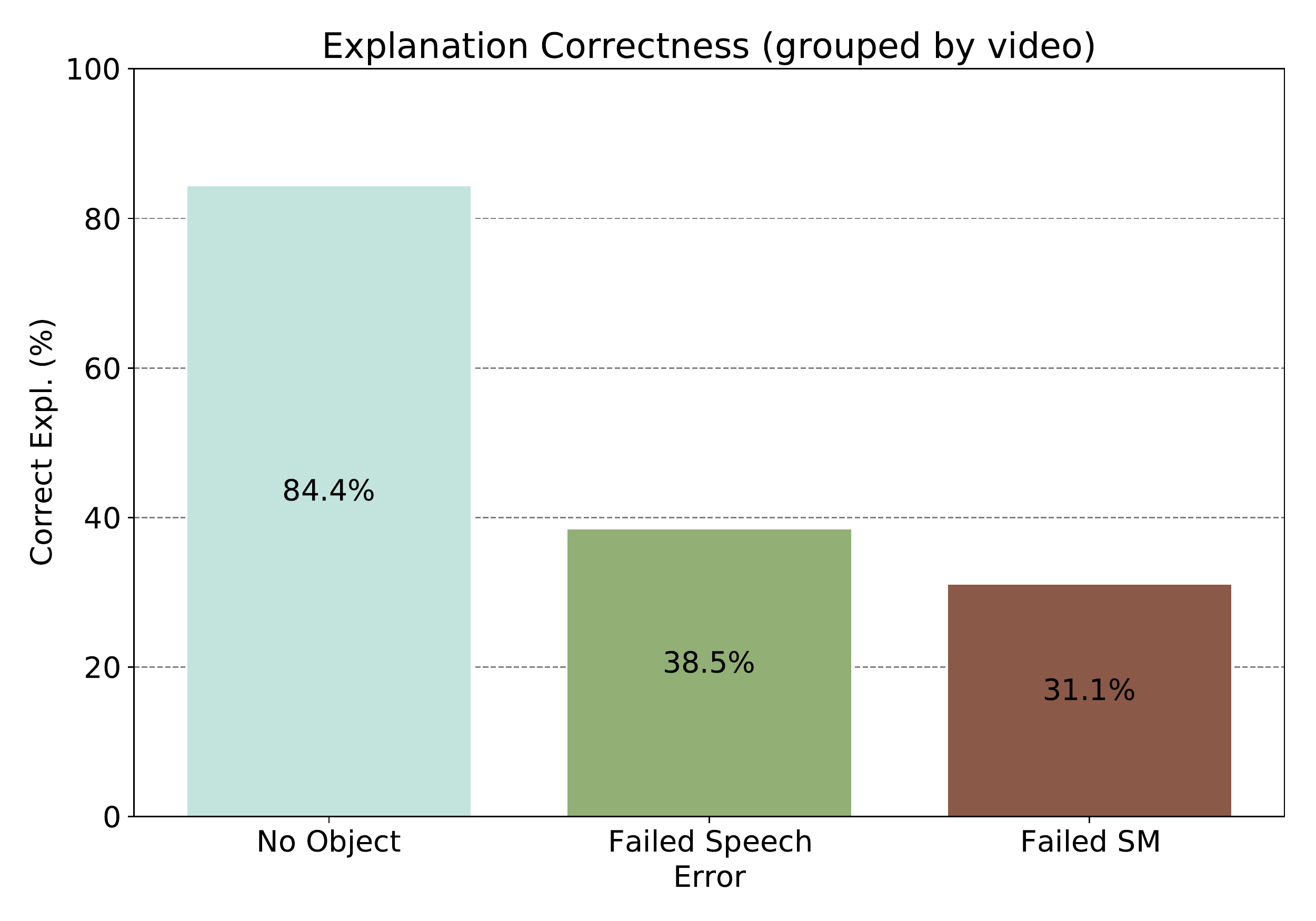}
    \caption{Amount of correct explanations per video.}
    \label{fig:observability}
\end{figure}

A key figure for this hypothesis was the number of correct explanations for each video (\cf~\cref{fig:observability}). The Kruskal-Wallis test indicated a significant difference between the videos (H-statistic = 81.1717, p-value = 0.000). The subsequent post-hoc Dunn test revealed that the \texttt{No Object} error was detected significantly more than the other errors (\cf~\cref{tab:observability}). While the \texttt{Failed SM} error was detected slightly less than the \texttt{Failed Speech} error, there was no significant difference.\\

\begin{table}
    \centering
    \caption{Kruskal-Wallis, with follow up Dunn test for the number of correct explanations per video. p-values $<$ 0.05 are highlighted in \textbf{bold}, p-values $<$ 0.1 are highlighted in \textit{italic}.}
    \begin{tabular}{l l S[table-format=1.4]}
    \toprule
     \multicolumn{3}{c}{\textbf{Kruskal-Wallis}(H-statistics = 81.1717, p-value = \textbf{0.0000})}\\\hline
     Condition1   & Condition2       & {p-value}            \\
    \hline\hline
    \rowcolor{gray!15}
    & Failed Speech & \textbf{0.0000} \\
    \rowcolor{gray!15}
    \multirow{-2}{*}{No Object} & Failed SM & \textbf{0.0000} \\
    Failed Speech & Failed SM & 0.2497 \\
    \end{tabular}

    \label{tab:observability}
\end{table}

\begin{figure}
    \centering
    \includegraphics[width=0.7\textwidth]{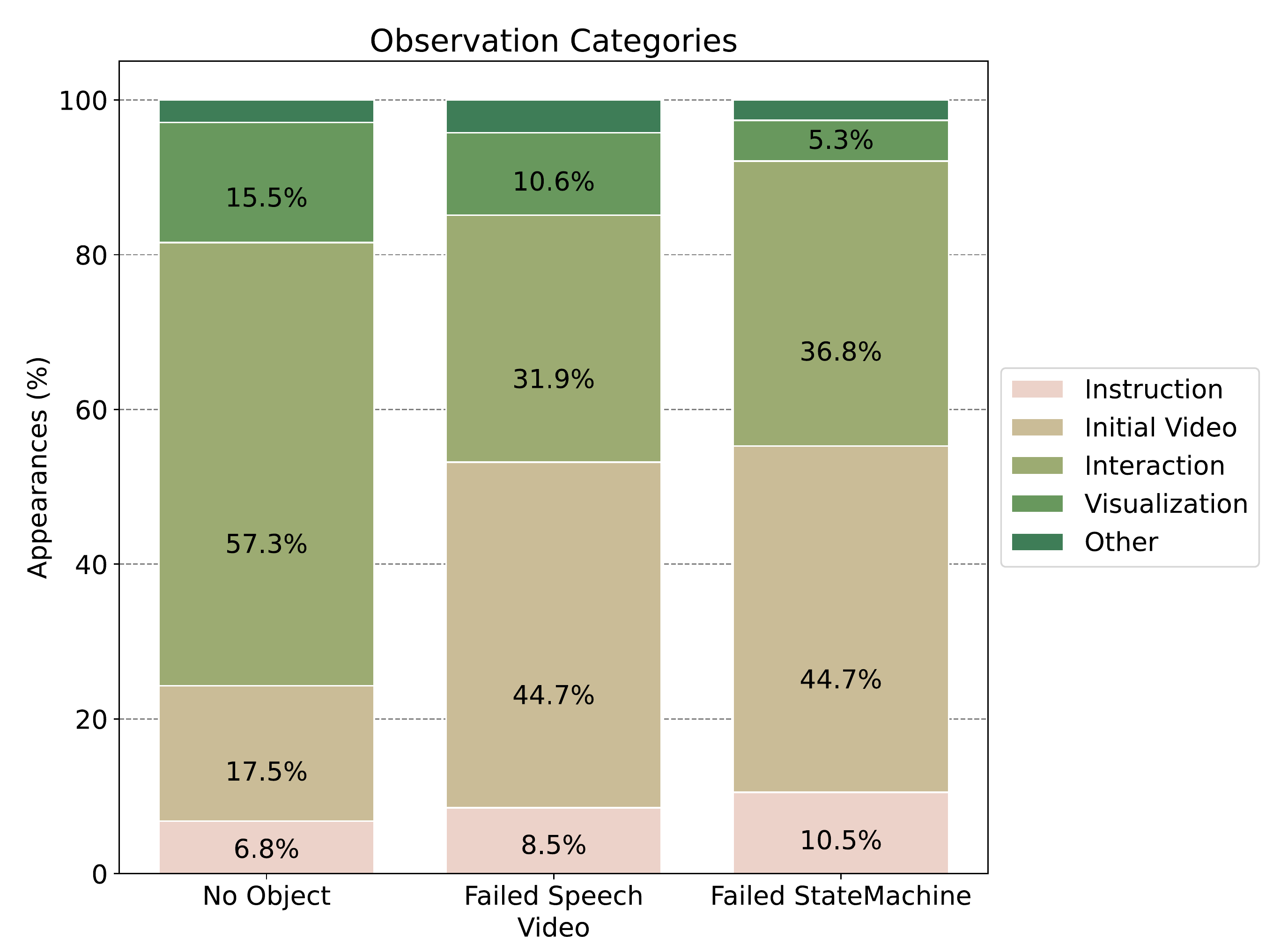}
    \caption{Categories of how the error was detected for correctly named error sources.}
    \label{fig:errorObserv}
\end{figure}

With regard to the third hypothesis, we also asked the participants after each video, how they recognized the error. To assign each answer to a category, we chose the following categories:
\begin{itemize}
    \item \texttt{Instruction}: The initial instruction video
    \item \texttt{Initial Video}: The initial working Human-Robot-Interaction video
    \item \texttt{Interaction}: The current Human-Robot-Interaction video
    \item \texttt{Visualization}: The additional visualization about the robot's internal states
    \item \texttt{Other}: Could not be categorized to one of the others
\end{itemize}
The answers can be seen in \cref{fig:errorObserv}. A Kruskal-Wallis test indicated significant differences for the \texttt{Initial Video} and the \texttt{Interaction} category between the videos. A post-hoc Dunn test revealed that the \texttt{Failed Speech} and \texttt{Failed SM} error videos were significantly more often detected by the \texttt{Initial Video} than the \texttt{No Object} error video. In contrast to this, the \texttt{No Object} error video was detected significantly more often in the current interaction video than the other two videos (\cf~\cref{tab:errorObserv}).

\begin{table}
    \centering
    \caption{Kruskal-Wallis, with follow up Dunn test for the number of correct explanations. Below the categories are the p-values of the Kruskal-Wallis test. Entries are p-values of the Dunn test. p-values $<$ 0.05 are highlighted in \textbf{bold}, p-values $<$ 0.1 are highlighted in \textit{italic}.}
    
    \begin{tabular}{l l S[table-format=1.4] S[table-format=1.4]}
    \toprule
    & & \multicolumn{2}{c}{{Error Observation}} \\
    \cmidrule{3-4}
    \multirow{2}{*}{Condition 1}    & \multirow{2}{*}{Condition 2} & {Initial Video}        & {Interaction} \\
                                    &                              & {\textbf{0.0003}}    & {\textbf{0.0061}} \\
    \midrule
    \rowcolor{gray!15}
                                    & \texttt{Failed Speech}                 & \textbf{0.0008}            & \textbf{0.004}\\
    \rowcolor{gray!15}
    \multirow{-2}{*}{\texttt{No Object}} & \texttt{Failed SM}            & \textbf{0.0017}            & \textbf{0.0314} \\
    \texttt{Failed Speech}               & \texttt{Failed SM}            & 0.9955            & 0.6517 \\
    \bottomrule
    \end{tabular}
    \label{tab:errorObserv}
\end{table}

\paragraph{ATI, Godspeed and System-Usability-Scale}
At the beginning of the survey, participants were asked to fill out the \emph{ATI} questionnaire \parencite{ATI}. Furthermore, the participants were asked to fill out parts of the \emph{godspeed} and the \emph{system-usability-scale (SUS)} questionnaire at the end of the survey. We only included the key figures \emph{anthropomorphism}, \emph{likeability} and \emph{perceived intelligence} from the \emph{godspeed} questionnaire.

To evaluate the influence of a better understanding of the robot towards the questionnaire scores, we grouped all participants by number of correctly explained erroneous videos (i.e. participants with zero correct explanations were in group \texttt{0}, those who could detect one in \texttt{1}, etc.). A one-way ANOVA showed significant differences for the \emph{anthropomorphism} and \emph{SUS} score. The follow up post-hoc tukey-HSD test revealed that participants who were able to explain all three errors had a significantly lower \emph{anthropomorphism} score than those who could not detect any errors, while the \emph{SUS} score was higher in contrast to the other participants with less correctly explained errors (\cf~\cref{tab:questionnaires}). Another finding was that the \emph{ATI} score, which reflects the technical affinity of the participants, had no influence on the result.

\begin{minipage}{0.5\textwidth}
    \centering
    \includegraphics[width=\textwidth]{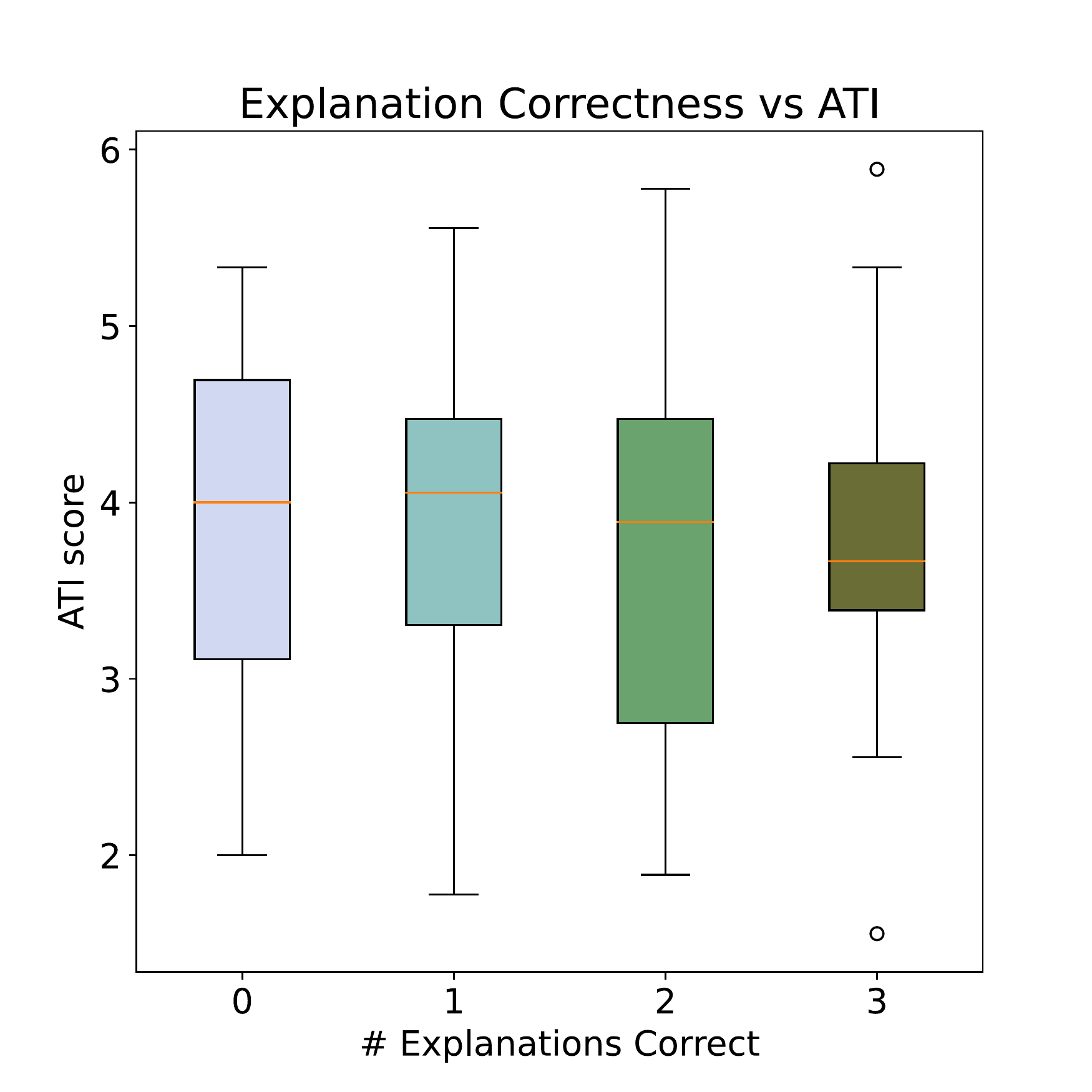}
    \captionof{figure}{Boxplot of the ATI scores grouped by number of correct explanations.}
    \label{fig:expl_ati}
\end{minipage}%
\vspace{1em}
\begin{minipage}{0.5\textwidth}
    \centering
    \includegraphics[width=\textwidth]{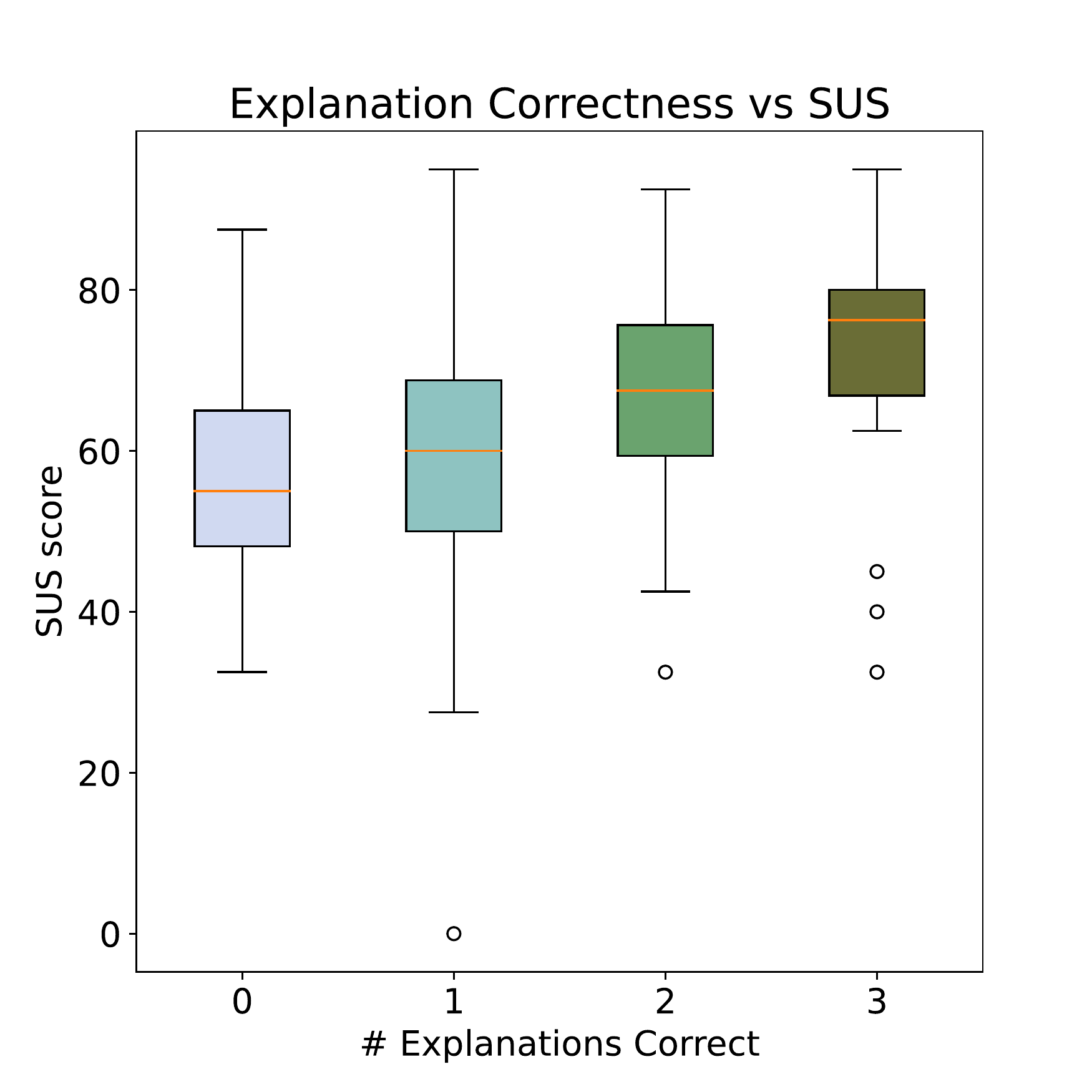}
    \captionof{figure}{Boxplot of the SUS scores grouped by number of correct explanations.}
    \label{fig:expl_sus}
\end{minipage}%

\begin{figure}
    \centering
    \includegraphics[width=\textwidth]{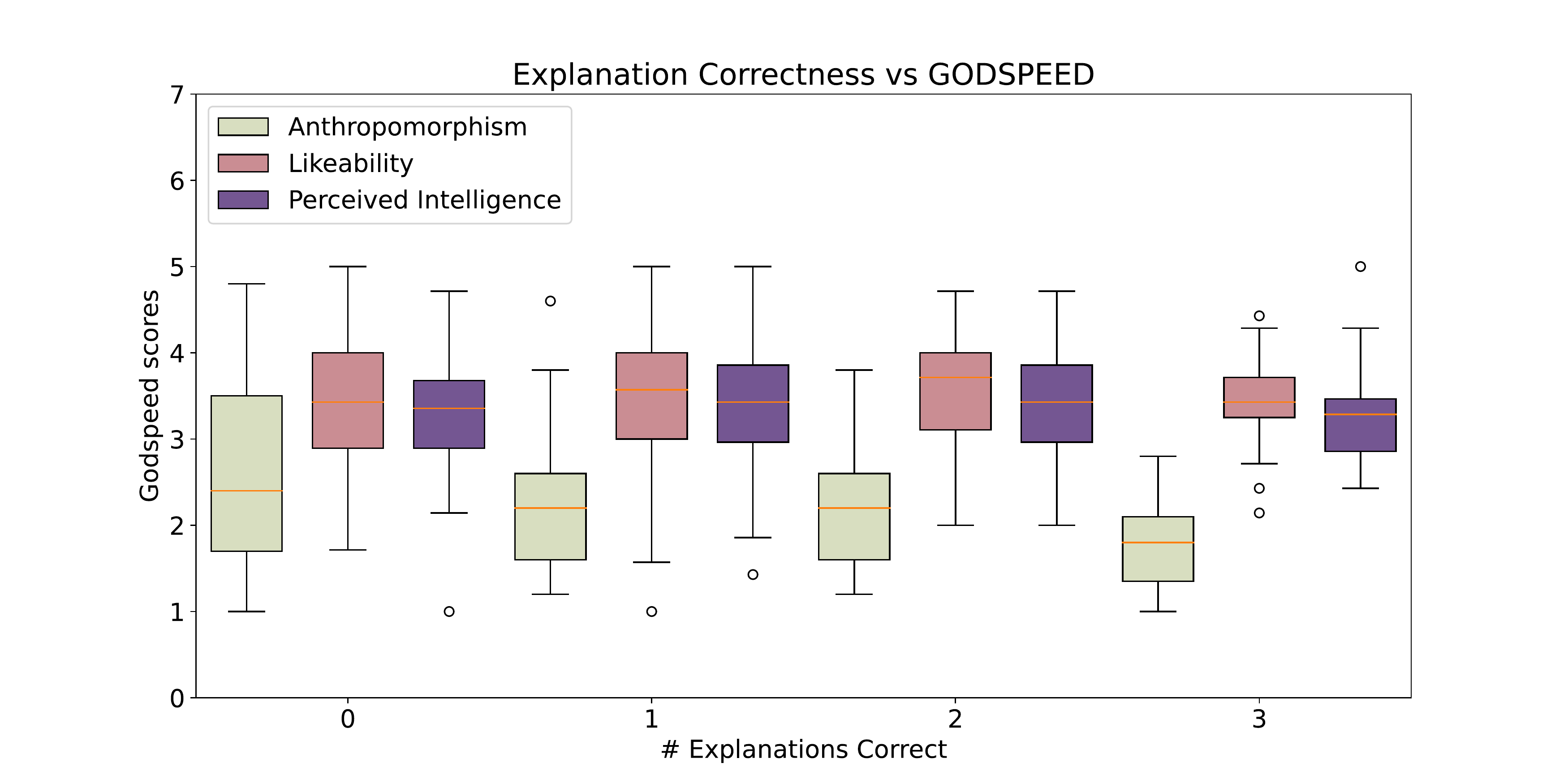}
    \caption{Boxplot of the GODSPEED scores grouped by number of correct explanations.}
    \label{fig:expl_god}
\end{figure}

\begin{table}
    \centering
    \caption{Contingency tests for the godspeed and SUS scores, grouped by number of correctly explained errors. Entries are: F-statistics (p-value). p-values $<$ 0.05 are highlighted in \textbf{bold}, p-values $<$ 0.1 are highlighted in \textit{italic}.}

    \begin{tabular}{c c S[table-format=1.4] S[table-format=1.4] S[table-format=1.4] S[table-format=1.4] c}
    \toprule
    \multicolumn{4}{l}{\textbf{ATI} (F-statistic = 0.7618, p-value = 0.5177)} \\
    \midrule
    \multicolumn{4}{l}{\textbf{Anthropomorphism} (F-statistic = 3.2010, p-value = \textbf{0.0259})} \\
    \midrule
    Condition 1    & Condition 2 & {meandiff}         & {p-adj}        & {lower}               & {upper} & {reject} \\
    \midrule
    \rowcolor{gray!15}
                                    & \texttt{1}                 & -0.4264 & 0.2856 & -1.0487 & 0.1959 & False\\
    \rowcolor{gray!15}
                                    & \texttt{2}                 & -0.3884 & 0.4253 & -1.0507 & 0.2739 & False \\
    \rowcolor{gray!15}
    \multirow{-3}{*}{\texttt{0}} & \texttt{3}            & -0.8071 & \textbf{0.0159} & -1.5022 & -0.1121 & True\\
    \multirow{2}{*}{\texttt{1}}   & \texttt{2}                 & 0.038  &  0.9 & -0.4264 & 0.5024 & False \\
                                    & \texttt{3}            & -0.3808 & 0.215 & -0.8908 & 0.1293 & False\\
    \rowcolor{gray!15}
    \texttt{2}               & \texttt{3}                 & -0.4187 & 0.2111 & -0.9769 & 0.1394 & False \\
    \midrule
    \multicolumn{4}{l}{\textbf{Likeability} (F-statistic = 0.3179, p-value = 0.8124)} \\
    \midrule
    \multicolumn{4}{l}{\textbf{Perceived Intelligence} (F-statistic = 0.5137, p-value = 0.6736)} \\
    \midrule
    \multicolumn{4}{l}{\textbf{SUS} (F-statistic = 4.1649, p-value = \textbf{0.0076})} \\
    \midrule
    Condition 1    & Condition 2 & {meandiff}         & {p-adj}        & {lower}               & {upper} & {reject} \\
    \midrule
    \rowcolor{gray!15}
                                    & \texttt{1}                 & 2.2596  &  0.9 & -10.5416 & 15.0609 & False\\
    \rowcolor{gray!15}
                                    & \texttt{2}                 & 8.75 & 0.3424 & -4.8734 & 22.3734 & False \\
    \rowcolor{gray!15}
    \multirow{-3}{*}{\texttt{0}} & \texttt{3}            & 14.7917 & \textbf{0.0396} &  0.4939 & 29.0894  & True\\
    \multirow{2}{*}{\texttt{1}}   & \texttt{2}                 & 6.4904 & 0.2928 & -3.0619 & 16.0427 & False \\
                                    & \texttt{3}            & 12.5321 & \textbf{0.0123} &  2.0404 & 23.0237  & True\\
    \rowcolor{gray!15}
    \texttt{2}               & \texttt{3}                 & 6.0417 & 0.5174  & -5.4388 & 17.5221 & False \\
    \bottomrule
    \end{tabular}
    \label{tab:questionnaires}
\end{table}

\section{Discussion}
\label{sec:discuss}
The results of the online survey, presented in \cref{sec:res}, provide good insights for our hypotheses. The \textbf{first hypothesis} stated that providing insights on the architecture of the robot will improve knowledge and understanding of such. Results from listing the hardware and software features show that architecture information indeed improve the understanding of what hardware features are used by the robot. While this is true for features that might not be well-known, like the \emph{microphone} or the \emph{pan-tilt unit}, mentions of features that are already known by the majority are not boosted. In contrast to the hardware features, the understanding of the software features could not be improved by architecture instructions. Instead a contrary observation could be made: the architecture condition mentioned less software features than the other conditions. This problem might have occurred based on how the instruction video was designed. A hardware feature was mentioned before its use on the software side was explained. Therefore, an overload of information could have led to only memorizing the hardware features. In addition, the video may have primed users so that they were only focused on information they remembered from the instructions.

Our \textbf{second hypothesis} stated that architecture insights help to recognize and explain errors in human-robot interactions, as these improve the mental model of the user about the robot. We tested this hypothesis with two different approaches regarding how architecture information is communicated. Our results were not able to support our hypothesis when information is presented in an instruction video. The presented architecture information does not improve the ability to explain errors in human-robot interaction. This might be due to the fact that although a user has an improved initial mental model of the robot, they lack insights regarding the current status of the robot and therefore cannot apply the architecture knowledge. Just as the instruction video, the visualization of the current internal state of the robot did not improve the ability to explain errors either. Instead, it actually seems to make it worse for some error types (i.e. \texttt{Failed Speech} and \texttt{Failed SM}). This problem might be explained by the quality and intuitiveness of our visualization, which might have caused an overload on information and its misinterpretation. Additionally, a visualization without further information on what it communicates allows much freedom for interpretation. This leads to situations where the visualization does not help at all, but confuses the user in a way that errors cannot be explained.

However, both approaches seem to complement each other such that the issues of either approach can be solved. The results show that in the fourth condition of showing an instruction video and providing a visualization, the ability to detect errors is significantly greater. By improving the mental model through the instruction video, the visualization is less confusing and its meaning can be more easily recognized. Additionally, the visualization helps to observe the learned architecture information throughout the human-robot interaction, which leads to an improved ability to detect and explain erroneous interactions.
 
Additional support for our hypothesis regarding the more appropriate shaped mental model (i.e., not equal to the mental model for another human) is given by comparison of the questionnaire scores for the number of correctly explained errors. The \emph{anthropomorphism} score is lower for participants who could explain all three errors. Furthermore, the \emph{SUS} score is higher for participants with more correctly explained errors. This suggests, that these participants ascribed less human-like characteristics to the robot, while rating the system as useful despite its apparent limitations. Furthermore, we compared the technical affinity of the participants who could explain the errors and those who could not. The results did not show any significant differences. Therefore, our strategies of communicating insights of the robot seem to be understandable for people with a technical affinity as well as for those without. This confirms that our approach is also applicable for naive users.

While we can see that architecture information improves the ability to explain errors in interaction and therefore improves the user's mental model, this ability also depends on the concept where the error occurs. The results from our \textbf{third hypothesis} showed that the error in the erroneous object recognition interaction could be explained much better than the speech recognition or the state machine. This can be primarily explained by the better observability and familiarity of the object recognition concept. More than half of the participants stated that they had observed the object recognition error in the interaction itself. The speech recognition and state machine errors, on the other hand, were detected by reference to the video of the previous correctly working interaction. We assume that this is due to the fact, that even with theoretical knowledge about technical concepts, users need to observe the correct explicit structure of the interaction to be able to identify an error.

These results suggest that it is reasonable to communicate technical concepts to users, in order to improve their mental model. Furthermore, we showed that the used concept has a major influence on the ability to detect an error. The concept of object recognition was designed in a way that people could use previous knowledge of QR-scanners. Based on this, we argue that designing robotic concepts should be geared to already established technical concepts. Together with communicating the concepts and increasing their observability through visualizations, an improved user's mental model about the robot can be formed. Although we showed that communicating technical concepts helps shaping a more appropriate mental model, we cannot make a general statement about the design of such information. With regard to the cognitive load of users, further studies should be carried out to evaluate how much information about a concept contributes to better understanding of such. Every concept has its own important factors that need to be communicated.\\

Our observations show that the concept of finite state machines, which is at the core of human-robot interactions, is neither familiar nor observable. Hence, further advances in that direction have to be made. It is reasonable to take into account the observations regarding pragmatic frames (\cf~\cref{sec:pf}). \Textcite{vollmer2016pragmatic} investigated the use of pragmatic frames in human-robot interactions. They showed that an interaction with a robot, where the user teaches the robot, is still not flexible enough for humans to intuitively interact with the system. We suggest that an alternative framework should allow for simple reasoning on the robot's side to communicate its internal processes. At the same time such a framework needs to be more flexible and adaptable in order for naive users, who have no experience with robotics, to intuitively interact with the system. In that way, the mental model develops while shaping the robot's behavior, based on the user's expectations. \Textcite{kaptein2017personalised} used a \emph{belief-desire-intention (BDI)} based agent to communicate the decision making process. While they showed that adults prefer explanations in terms of the robot's goals, we belief that this forms an incorrect mental model about the robot with unrealistic expectations regarding its cognitive abilities. Moreover, a \emph{BDI} architecture with its hierarchical structure might not be suitable to be taught by naive users, taking into account the potential complexity of such structures. \Textcite{saunders2015teach} developed a flexible system that can be personalized by the user. The decision making process is based on current sensor states, similar to reactive behaviors. Such a system is based on technical concepts but probably still simple enough for humans to understand. Thus, it could present a more intuitively understandable alternative to currently widely used state machine frameworks.

\section{Conclusion}
\label{sec:conclusion}
This paper deals with the question of how to improve users' mental models of robots in Human-Robot-Interactions. We investigated a new approach by communicating insights of the robot's architecture in two different ways. One approach tried to communicate technical concepts in an direct way, similar to a manual. The second approach was more indirect by visualizing the current internal states of the robot.

We evaluated the approaches in an online survey, where participants had to detect and further elaborate erroneous Human-Robot-Interactions. In addition to both approaches, we compared them to a baseline and also as combined approach.

The results showed that both approaches on its own doesn't help to explain erroneous situations. The visualization even worsened the ability by distracting the participants with too much information. In contrast to this, the combination of both approaches improved the ability to detect errors significantly. Furthermore, the ascribed anthropomorphic characteristics were lower, while the usability were rated higher.

Based on the results, we conclude that the current trend of social robotics, pretending that the robot has human-like abilities and emotions, might not be the optimal direction of research. Our society is confronted with technology in every days live for so long, that people are already able to deal with technical concepts. Our results suggest that, while improving the capabilities of robotic systems, the technical concepts of the robot with its limitations should be communicated to achieve common ground on the mental models of the robot and the user.

\section*{Acknowledgement}
This work was funded by the Honda Research Institute Europe, 63073 Offenbach am Main, Germany.

\printbibliography

\end{document}